# CONDITION MONITORING OF TRANSFORMER BUSHINGS USING COMPUTATIONAL INTELLIGENCE: FOCUS ON ATTRIBUTE REDUCTION

By

JOSHUA TSHIFHIWA MAUMELA

A dissertation submitted for the fulfilment of the requirements for the degree

MAGISTER INGENERIAE

In

ELECTRICAL AND ELECTRONICS ENGINEERING SCIENCE

At the

UNIVERSITY OF JOHANNESBURG

SUPERVISOR: Prof F.V Nelwamondo

CO-SUPERVISOR: Prof T Marwala

NOVEMBER 2013

# **Declaration**

I declare that this dissertation is a result of my own, unaided work, except where otherwise acknowledged. It is being submitted for the degree of Magister Ingeneriae at the University of Johannesburg. It has not been submitted before for any degree or examination in any other university.

Signed this ________ day __________________________ 20 ___

__________________________

Joshua Tshifhiwa Maumela

# **<u>Acknowledgments</u>**

Firstly I will like to thank God Almighty for making all this possible. I will then like to thank my Supervisors throughout this dissertation Professor Fulufhelo Vincent Nelwamondo and Professor Tshilidzi Marwala.

The constant guidance by Lindokuhle Mpanza gave me too much insight to what this work requested and for that I will always be grateful. It was also the words of encouragement and advice from Professor Bhekisipho Twala that also helped me get a clear vision and ideas of what I need to do.

I will always be grateful to my parents Jonas Ratshibvumo and Ndifelani Emily Maumela who forever supported me and encouraged me throughout my life with the endless support from my brother Dr Munaka Chris Maumela. To my brother Munaka, I will like to say your words of advice will always and forever remain in my heart; I will carry them with me wherever I go.

Thanks to the many family and friends who have always kept in touch to find out how my research was going. Your support will not go by unnoticed.

# Abstract

Dissolved Gas-in-oil analysis (DGA) is used to monitor the condition of bushings on large power transformers. There are different techniques used in determining the conditions from the data collected, but in this work the Artificial Intelligence techniques are investigated. This work investigates which gases in DGA are related to each other and which ones are important for making decisions. When the related and crucial gases are determined, the other gases are discarded thereby reducing the number of attributes in DGA. Hence a further investigation is done to see how these new datasets influence the performance of the classifiers used to classify the DGA of full attributes. The classifiers used in these experiments were Backpropagation Neural Networks (BPNN) and Support Vector Machines (SVM) whereas the Principal Component Analysis (PCA), Rough Set (RS), Incremental Granular Ranking (GR++) and Decision Trees (DT) were used to reduce the attributes of the dataset. The parameters used when training the BPNN and SVM classifiers are kept fixed to create a controlled test environment when investigating the effects of reducing the number of gases. This work further introduced a new classifier that can handle high dimension dataset and noisy dataset, Rough Neural Network (RNN). This classifier was tested when trained using the full dataset and how it is affected by reducing the number of gases used to train it. The results in these experiments showed that ethane and total combustible gases attributes are core attributes chosen by the four algorithms as gases needed for decision making. The average results of the classification performance showed that the reduction of attributes helps improve the performance of classifiers. Hence the science of transformer condition monitoring can be derived from studying the relations and patterns created by the different gases attributes in DGA. This statement is supported by the classification improvements where the RNN classifier had 99.7% classification accuracy when trained using the three attributes determined by the PCA.

## Table of Contents

# List of Figures

# List of Tables

# Acronyms and Symbols

**ANN** – Artificial Neural Networks

**BPNN** – Backpropagation Neural Network

**DGA** – Dissolved Gas-in-oil Analysis

**DT** – Decision Trees

**GrC** – Granular Computation

**GR++** - Incremental Granular Ranking

**IEEE** – Institution of Electrical and Electronic Engineering

**IEC –** International Electrotechnical Commission

**PCA** – Principal Component Analysis

**RNN** – Rough Neural Networks

**RS** - Rough Sets

**RST** – Rough Set Theory

**SVM** – Support Vector Machines

# 1 INTRODUCTION

## 1.1 Background Theory

Transformers are very large and critical electrical equipment [1]. The failure of power transformers may lead to a very large loss of money as most industrial production lines depend on them. Some industrial operations that highly depend on power transformers are power generation, mines, manufacturing plants, etc. There are various causes of a transformer failure, with some being factory defaults and majority being the faulty conditions. Factory defaults can be avoided by tests in the quality control labs before transformer is installed in the system.

Faulty conditions are however the major cause and can only be determined through monitoring the conditions of the transformer. If these conditions are not monitored, the transformer might reach the end of its life without being noticed by the operator or fail prematurely. Such instances lead to condition monitoring of electrical equipment being of importance and therefore a subject of research. By monitoring the operating conditions the operators do not have to only depend on the quality and the word of the supplier to know the lifespan of their equipment.

Mechanical equipment wears out due to vibrations and rust mostly. Different angles of vibrations, be it in dynamic or static equipment, lead to wearing of mechanical structures. Electrical equipment is mostly affected by voltage and current which may also lead to corrosion. In electromechanical equipment a combination of electrical and mechanical factors that lead to wearing can be experienced. Power transformers are an example of equipment that can wear off due to vibrations and electrical factors.

To monitor the conditions of a transformer, there are few techniques that have been developed, these techniques analyse electrical, thermal, vibration and dissolved gas analysis (DGA) properties. The dissolved gas analysis method has been investigated and studies have shown that the DGA can be used to determine the electrical and thermal faults in the transformers. It was not until the early 1960s/ late 1970s that J Morgan [2] et al adopted the technique of analysing of gases from petroleum products using chromatography in the power transformers. This technique got well known and successfully applied in many international engineering companies. It has thus become a standard practice in power industries to use this technique, since power transformers have oil used for cooling and insulation [2-5].

The advantage of this technique is that the oil samples can be extracted from most equipment without interrupting production. There are arrays of physical, chemical and electrical tests conducted on the extracted oil, but these tests can only give information about the condition of the oil and not so much information about the operating conditions of the transformer itself.

Thus in order to diagnose the operating conditions of these machines, DGA is the test to be performed. From the DGA results the operators can know the different faults that occurred and can also perform a root cause analysis to find out what happened. These different fault types include partial discharges, thermal faults, corrosion, etc. These different faults cause formation of different gases and different gas concentration in the oil, thus from the analysis of these gases the information about the type of faults can be revealed.

There are five widely used conventional methods for analysing and interpreting the condition of the transformer from the acquired gases in DGA:

- Doeneberg's Ratio Method
- Roger's Ratio Method
- Basic Gas Ratio Method
- Number Code Ratio Method
- Duval's Triangle Method

The first four methods use the ratio where the key gases are used to divide the other DGA gases to interpret the type of faults experienced by the transformer. The Duval Triangle method uses percentage as opposed to the first four methods. The Duval's

Triangle sums up the $C_2H_2, CH_4,$ & $C_2H_4$ to be the denominator and calculates each portion by taking $C_2H_2, CH_4,$ & $C_2H_4$ to be a member separately.

The presence of Hydrogen in the oil indicates the partial discharges, heating and arcing. The presence of methane, ethane and ethylene indicate the "Hot Metal" gases which are believed to be caused by circulating currents [6], while Acetylene indicates arcing and carbon oxides indicate cellulose insulation degradation.

## 1.2 Literature Review and Problem Definition

There are various methods used to monitor the condition of transformers that have oil for insulation and lubrication, these methods include vibration analysis, electrical properties analysis, thermal and dissolved gas-in-oil analysis (DGA). The latter is the one which is investigated in this work. DGA analysis is used because through it the different electrical properties can be determined and used to diagnose the faults within the transformer [2-7].

Since its first introduction, DGA has shown significant growth with more laboratory methods being developed and more online analysis methods being developed [7]. The successful application of DGA in transformers has resulted in the applications of DGA condition assessment techniques in other equipment that have oil and cellulose as insulation.

To monitor the condition of transformers, the gas is extracted from the oil in the transformer for analysis [2-7]. The extracted gasses include Hydrogen ($H_2$), Methane ($CH_4$), Ethane ($C_2H_2$), Ethylene ($C_2H_4$), Acetylene ($C_2H_6$), Oxygen ($O_2$), Nitrogen ($N_2$), Carbon Dioxide ($CO_2$) and Carbon Monoxide ($CO$). In some applications all combustible gases are added together to form another attribute called Total Combustible Gases (TCG). These gases are formed within the oil because of a particular electrical fault [2]. These characteristics are the ones used to classify the gas presence into different fault classes.

To assist with the analysis of these gases in monitoring the transformer conditions, IEEE [31] and IEC [32] have developed standards that are followed during the analysis. Through these standards researchers have developed various methods to analyse the gases-in-oil. These methods include Doeneberg's Ratio method [7], Roger's Ratio method [7], Basic Gas Ration method [7], Number Code Ratio method [7] and Duval's Triangle method [7]. These methods require experts to interpret them. This gives rise to the need for autonomous methods to analyse the data; and such methods include the use of computational intelligence methods for DGA analysis [8-30].

The adoption of computational intelligence methods addresses problems such as age, cumulative gas residues from previous events, gas build up in time and transformers of different families by learning the patterns from previously diagnosed transformers. The learnt patterns help to classify new unseen problem accurately and autonomously. The other reason why computational intelligence algorithms are best for interpreting these gases is that, the operators may not always know the exact applied voltage that caused faults and the duration of the arc. These two determine the gas concentration level of gases non-linearly and linearly respectively. These algorithms solve this problem by identifying and learning previous patterns.

Artificial Neural Network (ANN), Support Vector Machines (SVM) and Rough Sets Theory (RST) are amongst the computational intelligence techniques that have been investigated in the application of analysing dissolved gas-in-oil. Dhlamini et al. [8-10] investigated using artificial neural networks and support vector machines in the analysis of DGA. Mpanza [11] investigated using rough sets for analysis because of their ability to have easily interpretable results. Zhang et al. [12] continued on the work done in [8] to show that the insertion and deletion of CO as an attribute to be trained using ANN influences the classification accuracy.

In [10-30] the key gasses were used to train the Neural Networks and Support Vector Machines where the key gases were used to create a set of three attributes using the IEC Standards. This standard involved using the key gases to calculate the gas ratio which was then used in diagnosis.

Venkatasami et al. [14] investigated using IEC gas ratio methods to create the input attributes against the GECB method and finally against using the key gases chosen as optimal $H_2, CH_4, C_2H_2, C_2H_4, C_2H_6$ by [13]. The results of [14] showed that the ANN trained using the key gases had higher accuracy (82% compared to 78.67% and 74.66% where gas ratios were used as input). The ANN configuration parameters were kept constant during the comparative experiments. This created same condition environment to find the benchmark for the comparative study. This work followed a similar approach. The influence of each dataset preparation becomes apparent as the training time and classification accuracy differ when training data is used to train the same neural network. This work used a similar approach because it gives a clear indication of the influence of each dataset on training the same classifier.

Most of these works used IEC 60567 Standards [32] and not considering the usage of Carbon Monoxide and Carbon Dioxide gases as attributes in training, this meant that they were assuming that there is no cellulose degradation [1-2,7] in the transformers since they did not exclude it through data preprocessing. In [10] Dhlamini et al. showed that Carbon Monoxide can be influential in the diagnosis of the transformer bushings in the presence of an extra attribute, oil level but Carbon Dioxide is least influential. Researchers in these works [12-29] were using a small dataset of samples ranging from 100 to 400 observation instances. The work done by [12] shows what CO and CO2 have an influence in the performance of the classifiers but these experiments were done in the presence of only five extra gases $H_2, CH_4, C_2H_2, C_2H_4, C_2H_6$ and the presence of $N_2, O_2$ and Total Combustible Dissolved Gasses (TCDG) was ignored.

In this work the data preprocessing stage where the input attributes are chosen from the dataset is proposed. The proposed methods chooses the input attributes by identifying the patterns, correlation and dependency of attributes in making a decision in the presence of all dissolved gases in the dataset and with a large dataset of 20000-40000 observation samples. A large dataset is chosen to reduce the biasness which might result from the data patterns. The gas attributes are used during analysis instead of gas ratios because in [25] it was shown that training the SVM classifier with the key gases yielded better results than when gas ratios are used. The proposed methods to reduce the attributes of the bushings DGA dataset are Principal Component Analysis (PCA), Rough Set Theory (RST), Granular Computation (GrC) and Decision Tree (DT).

The Rough Set Algorithm to be investigated in this work chooses the attributes by evaluating the dependency of the decision in the training data on each attribute, using the Degree of Dependency algorithm, while the PCA algorithm chooses the attribute depending on the correlation of the different attributes. The GrC's algorithm uses the same algorithm as the RS algorithm but in an increasing dataset hence the Incremental Granular Ranking (GR++) algorithm was used [33-37]. Decision Tree algorithm is also investigated the dependency of decision attribute on conditional attributes. The resultant attributes of each preprocessing technique is used to train the ANN and SVM.

Even though these methods can be used in fault diagnostic application in this work they are not be used for that but for identifying the minimum number of attributes that can be used to train the classifiers. These attributes must train the classifiers without compromising the performance of the classifier but improve it in terms of training speed or classification accuracy. The classifiers used in this work are limited only to fault detection and not used in diagnosis. Thus this work does not investigate transformer bushings fault diagnosis but transformer bushings fault detection.

To address the classification of vague dataset and nonlinear dataset, this work further introduces the application of Rough Neural Networks (RNN) which was introduced by Lingras [38-42]. Rough Neural Networks had never been applied in fault detection applications of bushings DGA data in accordance with the author's knowledge. RNN are hybrid of rough sets and neural networks developed because RS are good with classification of high dimensional dataset but sensitive to noise whereas neural networks are robust to noise but weak in processing high dimensional data. As a result of the hybrid, RNN exhibit the good qualities of RS and ANN. The data preprocessed using the solutions proposed in this work was used to also train the performance of the RNN to benchmark its performance against the other SVM and the ANN classifiers.

## 1.3 Project Focus and Methodology

This work focuses on reducing the attributes of the dataset used to detect faults in the transformer bushings where the original attributes of the DGA dataset are ten. The preprocessing techniques are used to reduce attributes by learning the patterns and determining the dependency of attributes on each other in the training dataset. The chosen attributes are then used to train the SVM and the Backpropagation Neural Network classifiers (BPNN).

Thus the research procedure in this work is summarised as follows:
- Observe the classification performance for classifiers trained using full unpreprocessed dataset.
- Determine the correlation between the DGA gases and dependency of decision attribute on DGA gases and thus reduce number of gas attributes.
- Use data with high correlation and high dependency to classify conditions of bushings using DGA data.
- Introduce a classifier robust to noisy data, handles high dimension dataset and can classify non-linear dataset.
- Observe the performance of the newly introduced classifier when trained using dataset with less attributes.

The methods used to reduce the attributes of the DGA dataset are Principal Component Analysis, Rough Set, Incremental Granular Ranking and Decision Trees. The limitations and assumptions considered in this work are as follows:
- Classifiers are not used for fault diagnosis but for fault detection, i.e. this work does not classify the fault type.
- All fault types are considered as faults of equal magnitude or importance.
- Incomplete data was removed from the dataset by deleting all the rows with missing column values.
- Preprocessors are not used for classification of faults.
- Data preprocessing and classification are done independently with the data preprocessing done first.

- The attributes determined by each preprocessor in the data preprocessing stage were the ones used to train all classifiers, i.e. only one set of dataset for each preprocessor is used throughout this work.
- The data used in this work were collected from transformer bushings from South Africa. Results may differ for other countries due to environmental conditions.

This research then focuses further on the introduction of a new classifier in faults diagnosis, RNN. The performance of the RNN is then be compared to the performance of the BPNN and SVM using the same set of original dataset from DGA used to train the BPNN and SVM.

The research further investigates the classification performance of the RNN when trained using the dataset obtained from RS, PCA, GR++ and Decision Tree (DT) algorithms. The project focus can thus be pointed out below as follows:
- Introduction of new classifier, the Rough Neural Network(RNN)
- Compare RNN classifier's performance with BPNN and SVM classifiers.
- Determine the correlation between the gasses and reduce the attributes
- Determine the dependency of decision attributes to the gasses and remove the attributes that the decision does not depend on.
- Investigate the effect of training classifiers using the dataset with few gas attributes.

This work is divided into two main parts as indicated in Figure 1 and Figure 2. The first part is the investigation of data preprocessing techniques in improving classifier performance as shown in Figure 1. The second part is the investigation of RNN as a classifier in transformer bushings fault analysis as shown in Figure 2.

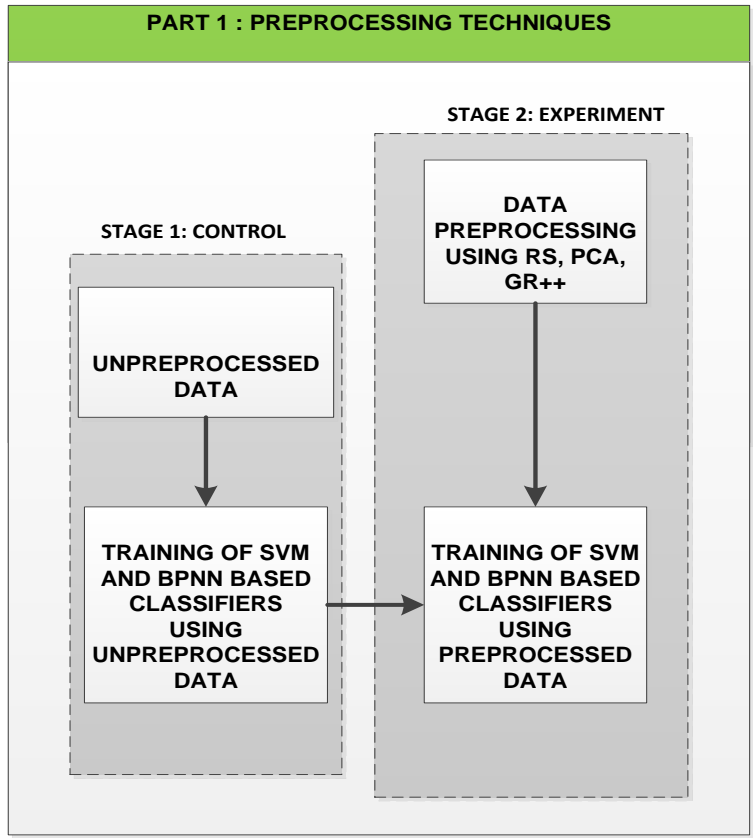

**Figure 1: Stages of the investigation of the preprocessing techniques**

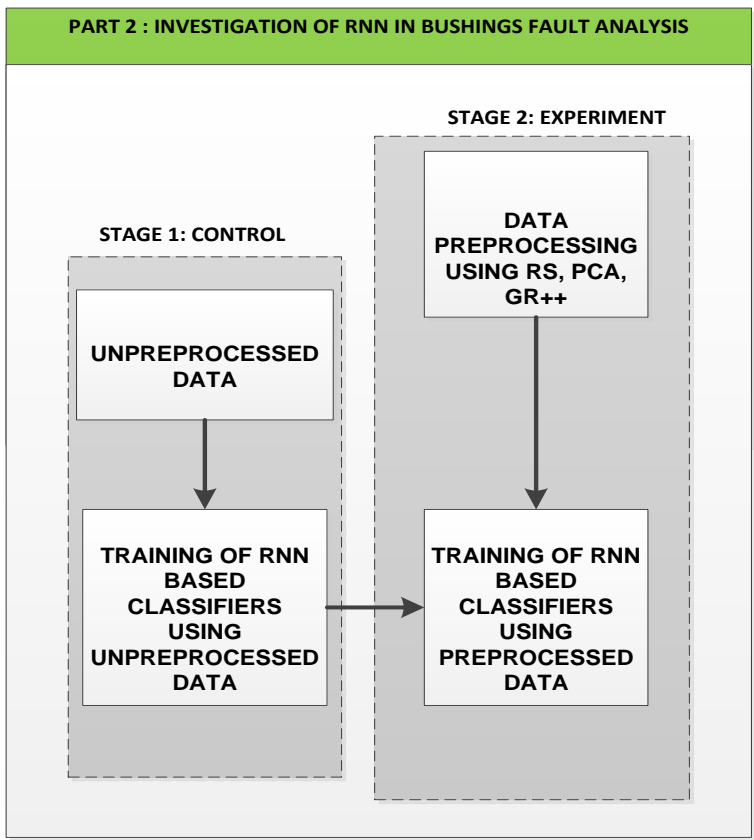

**Figure 2: Stages of the investigation of RNN in bushings fault analysis**

In the first part of this work, the processing techniques for DGA gases are investigated in two stages. In Stage 1, the Control, the SVM and BPNN classifiers are trained using unpreprocessed dataset. The best parameters used for training the classifiers when using unpreprocessed data are used as fixed control parameters for the experiments where the classifiers are trained using dataset with few attributes.

Stage 2, the Experiment, investigates using PCA, RS, GR++ and DT as data preprocessors. The correlation between gases and dependency of decision attribute on these gases is determined. Each one of these preprocessing techniques processes out a dataset with the best attributes as its output. This means that there are three datasets at the end of this stage.

The performance of the BPNN and SVM based classifiers when trained using the dataset with the best attributes according to the results of the data preprocessing experiment. The BPNN and SVM are then trained using the dataset from the PCA, RS, GR++ and DT algorithms. The performance of classifiers when trained using each dataset is then compared noting the positive and negative effects that each preprocessing technique has on either of these classifiers. Stage 2 of this work tests the dataset chosen by each preprocessing technique. The dataset chosen by the PCA, RS, GR++ and DT preprocessors is tested.

Part 2 of this work investigates the introduction of RNN in analysing faults in transformer bushings DGA data. This part is also done in two stages. Stage 1, the Control, investigates the performance of RNN in analysis of raw unpreprocessed data of the bushings fault data and compares the results with the performance of the BPNN and SVM classifiers trained using the same unpreprocessed dataset as in "Part1-Stage 1".

Stage 2, the Experiment, investigates the performance improvements of the RNN classifier when using the dataset from data preprocessors obtained in "Part1 – Stage 2". The results of the RNN performance when using dataset from PCA, RS, GR++

and DT is compared against the data obtained from "Part 1 – Stage 2" when the BPNN and SVM were trained using the same results.

## 1.4 Document Overview

Chapter 2

- Three methods used to reduce the dataset attributes are discussed
- Theory of each method is discussed in full detail
- The experimental setup and methodology followed in implementing each method is discussed
- Obtained results are analysed and discussed in this chapter.

Chapter 3

- Classification methods used to classify the data obtained in Chapter 3 are discussed
- Theory of each of these classifiers is discussed
- The experimental setup and methodology followed is discussed
- Results are discussed from each experiment
- Methods discussed in Chapter 3 are discussed in terms of their effects on each classifier.

Chapter 4

- A new classifier is introduced for DGA bushings fault detection applications
- The classifier is benchmarked against the classifiers investigated in chapter 4
- The effects of methods used in Chapter 3 are also tested in this chapter
- Results are then evaluated and compared

Chapter 5

- This chapter analyses the whole work
- The results obtained from Chapter 4 and Chapter 5 are compared against each other and the overall conclusion of the whole work is drawn

Chapter 6

- This chapter suggests the future work that is to build up on this current work.

# 2 ATTRIBUTE REDUCTION OF DISSOLVE GAS-IN-OIL ANALYSIS DATASET

## 2.1 Principal Component Analysis

### 2.1.1 Mathematical Theory

Principal component analysis has been used in many applications, of which most of them were in attribute reduction, data compression and some in pattern recognition. In most applications PCA is used in attribute reduction of multi-dimensional dataset. In [43-46] more work where PCA has been used successfully for data attributes reduction is discussed.

This work follows nine steps when using Principal Component Analysis to reduce the attributes of the dataset.
The steps are explained below with their mathematical equations.

   A. *__Data Preparations__*

   The dataset used in this work has ten conditional attributes and one decision attribute. The problem addressed in this work is a two class decision problem where the transformer bushing is either faulty or healthy. The condition attributes represent the different values of the gases found in oil which can be used to determine the condition of the transformer bushings.

**Table 1: Example of the Dataset Used in this Work**

| Acetylene | Carbon dioxide | Carbon monoxide | Ethane | Ethylene | Hydrogen | Methane | Nitrogen | Oxygen | Total Combustible Gases | Decision |
|---|---|---|---|---|---|---|---|---|---|---|
| 43 | 242 | 57 | 294 | 2055 | 1175 | 1882 | 12233 | 4060 | 5506 | 0 |
| 12.7 | 90 | 1491 | 0 | 89 | 1616335 | 297 | 31 | 52013 | 1618224.7 | 0 |
| 3479 | 898 | 61 | 1582 | 5409 | 206 | 548 | 41043 | 16462 | 11285 | 0 |
| 366 | 4745 | 1006 | 345 | 2358 | 558 | 1291 | 48346 | 8379 | 5924 | 0 |
| 1878 | 1961 | 1448 | 87 | 1198 | 1827 | 669 | 38756 | 15154 | 7107 | 0 |
| 10 | 2040 | 207 | 14 | 95 | 25 | 38 | 51405 | 17312 | 389 | 1 |
| 13 | 2336 | 252 | 28 | 204 | 56 | 88 | 56213 | 18532 | 641 | 1 |
| 8 | 1848 | 213 | 20 | 141 | 29 | 52 | 48833 | 16483 | 463 | 1 |
| 14 | 2528 | 340 | 56 | 404 | 115 | 170 | 56275 | 17456 | 1099 | 1 |

The condition attributes are presented in columns while the gases values for different bushings samples is presented in the rows. The decision attribute has two classes, 1 and 0. The classes represent a transformer that had been classified healthy and faulty respectively. Since PCA is an unsupervised learning method, the decision attribute is ignored when using PCA to reduce attributes.

### B. *<u>Find the Mean of the Dataset</u>*

The first major step in data analysis through PCA is to find the mean of each attribute in the dataset.

Let the dataset be $X$ of $n \times m$ dimensions where $n$ is the number of rows and $m$ is the number of columns.
Then the mean is defined in Equation (1)

$$\bar{X}_m = \sum_{i=1}^{n} \frac{X_{im}}{n} \qquad \text{Equation (1)}$$

Now that we have the average value of the columns in the dataset we can determine the average distances at which the data values of each attribute are away from the mean of that particular attribute. This is done in the next step.

C. **_Find the Standard Deviation of the Dataset_**

The standard deviation is the average distance of the data values away from the mean and this is obtained by computing the linear combination of the squared distance between the data values and the mean. The total sum obtained from the linear combination is then divided by $n$ thus finding the average. To complete the standard deviation calculation a positive square root of the average distance is taken.
This is mathematically represented as in Equation (2)

$$S_m = \sqrt{\frac{\sum_{i=1}^{n}(X_{im} - \bar{X}_m)^2}{n}} \qquad \text{Equation (2)}$$

The next step is the most important step in PCA.

D. **_Find the Zero-Mean of the Dataset_**

The data values in each attribute in this work vary in ranges. The attributes whose values have a large range may look more important than those of lower range values. This is a potential problem in PCA because more weight is given to them. To address this problem a standardisation method is used of which PCA used the zero-mean to achieve standardised data.

$$[Zero - Mean]_{im} = \frac{X_{im} - \bar{X}_m}{S_m} \quad \text{Equation (3)}$$

The next step is to determine the degree at which the attributes depend on each other by determining their covariance because correlation is the core of PCA.

E. **_Determine the Covariance of the Zero-Mean Data_**

To explain covariance the concept of variance which is the measure of how data spreads for each attribute is explained first. The variance is simply a squared standard deviation.

$$S_m^{\ 2} = \frac{\sum_{i=1}^{n}(X_{im} - \bar{X}_m)^2}{n} \quad \text{Equation (4)}$$

Covariance is thus the measure of how data spreads in an attribute as a result of other attributes. It is necessary to use the covariance instead of the variance in this work because PCA analyses the relationship between different attributes.

Assume we have attributes $X_{n1}$ and $X_{n2}$; then the covariance is determined using the following equation.

$$Cov(X_{n1}, X_{n2}) = \frac{\sum_{i=1}^{n}(X_{i1} - \bar{X}_1)(X_{i2} - \bar{X}_2)}{n} \quad \text{Equation (5)}$$

This means that the data spread of attribute $X_{n1}$ is measured against the data spread of $X_{n2}$.

In a multi-dimensional case, the calculation of covariance gets a bit tricky since the spread of data in each attribute must be calculated in relation to each attribute. For example, in a three dimensional dataset problem $(X_{n1}, X_{n2}, X_{n3})$ the covariance is determined between

$$\{(X_{n1}, X_{n2}) \quad (X_{n1}, X_{n3}) \quad (X_{n2}, X_{n3})\}$$
Equation (6)

The best way to calculate the covariance of a multi-dimensional dataset is to calculate all possible covariance combinations and store the values in matrix.

Covariance matrix of $m - dimension$ dataset is

$$C^{m \times m} = \left(C_{ij}, C_{ij} = cov(Dim_j, Dim_j)\right)$$
Equation (7)

$where\ Dim_x$ is the $x^{th}$ dimension.

There are three important characteristics of the covariance which can be used to interpret the covariance matrix.

The values are not as important as the signs before them.

i. <u>Positive Numbers</u>
This shows that the relationship between the attributes is of direct proportionality. This means that an increase in one attribute influences the increase in the other one as well.

ii. <u>Negative Numbers</u>
This shows that the relationship between the attributes is of inverse proportionality. This means that an increase in one attribute influences a decrease in the other attribute.

iii. <u>Zero</u>

This shows that there is no relationship between the two attributes. This means that the attributes are independent of each other.

**F. <u>Find Eigenvectors from the Covariance</u>**

Eigenvectors arise from the transformation matrix theory. Eigenvectors can only be determined in a square matrix but not every square matrix has eigenvectors.

In a $m \times m$ matrix the total number of eigenvectors is always equal the total number of $m$ columns. The most important characteristic of the eigenvectors used by the PCA is that all eigenvectors are orthogonal to each other.

The orthogonal property makes it possible for PCA to express the data points of the original dataset around the eigenvectors instead of the original axis.

Thus the eigenvectors are used to map new multi-dimension axis on which the data points is expressed around.

In this work, the eigenvectors are determined in unit length so that all eigenvectors have unit length. After determining the eigenvectors, the eigenvalue associated with these eigenvectors are determined.

**G. <u>Find Eigenvalues Associated with the Eigenvectors.</u>**

**Definition 2.1**: *If $A$ is a $n \times n$ matrix, then a non-zero vector $x$ in $\mathbb{R}^n$ is called an eigenvector of $A$ if $Ax$ is a scalar multiple of $x$; that is if*

$$Ax = \lambda x \qquad \text{Equation (8)}$$

*for some scalar $\lambda$. The scalar $\lambda$ is called an eigenvalue of $A$ and $x$ is said to be an eigenvector of $A$ of the corresponding eigenvalue.*

The following equation is used to determine the th eigenvalues in a matrix

$$Ax = \lambda I x \qquad \text{Equation (9)}$$

$$(\lambda I - A)x = 0 \qquad \text{Equation (10)}$$

In order for the solution to be eigenvalues they must satisfy the characteristic equation.

$$det(\lambda I - A) = 0 \qquad \text{Equation (11)}$$

The characteristic equation for high dimensional problems of $n$ dimensions is always a polynomial p in $\lambda$ and it's called a characteristic polynomial.

$$P(\lambda) = \det(\lambda I - A) = \lambda^n + c_1 \lambda^{n-1} + \cdots c_n \qquad \text{Equation (12)}$$

Eigenvalues are used in PCA to determine the principal components.

### H. <u>Determining Principal Components</u>

The principal components of the dataset are the eigenvectors with the highest eigenvalues. In this work the eigenvectors are ordered in a way of highest to lowest eigenvalue. This makes it easy to analyse and chose the desired principal components.

There are three methods used for selecting the best principal components from the eigenvalues. These methods are:

i. Eigenvalue-One criterion

In this method all eigenvectors whose eigenvalue is more than one are kept.

ii. The Scree Test

This method plots the eigenvectors against their eigenvalues and the points with big gaps is called the break points and all eigenvectors before the break point are kept.

iii. Variance Proportion

This method retains eigenvectors whose proportion of the eigenvalue over the total eigenvalue is higher. This method uses cumulative percentage method where the retained eigenvectors are those whose proportion percentage has reached a certain defined threshold.

This work used the variance proportion method. The proportion was calculated in percentage using the following equation.

$$Proportion = \frac{Eigencalue\ of\ Component\ of\ Interest}{Total\ Eigenvalue\ of\ the\ Correlation\ Matrix} \quad \text{Equation (13)}$$

After selecting the principal components, the vector containing the eigenvectors can be presented as follows.

$$FeautureVector = (eig_1 \quad eig_2 \quad eig_3 \quad \cdots \quad eig_p) \quad \text{Equation (14)}$$

Where p is the first kept eigenvectors thereafter called principal components.

I. Construct New Dataset

To construct the final dataset with few attributes the principal components are multiplied with the zero-mean dataset.

$$Final\ Dataset = FeatureVector^T \times (ZeroMean\ Dataset)^T \quad \text{Equation (15)}$$

The final dataset has p-dimensions.

$$where\ P < m$$

### 2.1.2 Methodology and Experimental Setup

This section shows the algorithm followed to design the Principal Component Analysis used in this work. Algorithm 2.1 shows how PCA algorithm was used to reduce attributes of the dissolved gases-in-oil dataset.

### *2.1.3 Experimental Setup*

MATLAB® was used to develop the principal component algorithm used to reduce the attributes. The raw DGA dataset is standardised by finding the zero-mean of the dataset. The final dataset constructed from the PCA algorithm was used to train the classifiers. The final dataset is the final output of the PCA and the values are standardised in respect to the zero-mean.

### 2.1.4 Results and Discussion

The original dataset had ten conditional attributes, of which in this experiment, PCA was used to reduce them. Three principal components were chosen as the components to represent the DGA dataset using the PCA. The three attributes indicated the data points that best presented the rest of the attributes.

| Algorithm 2.1 - Principal Component Analysis |
|---|

**Input:** DGA Dataset with 11 Attributes

**Output:** Dataset with Reduced Attributes

**Begin:**

Find the Mean of the Dataset

$$\bar{X} = \sum_{i=1}^{n} \frac{X_{im}}{n}$$

Find the Standard Deviation

$$S = \sqrt{\frac{\sum_{i=1}^{n}(X_{im} - \bar{X})^2}{n}}$$

Find the Zero-Mean

$$X - \bar{X}$$

Find the Covariance

$$C^{m \times m} = \left(C_{ij}, C_{ij} = cov(Dim_j, Dim_j)\right)$$

Find the Eigenvectors and Eigenvalues of Covariance

$$Ax = \lambda I x$$

Multiply Eigenvalue with Zero-Mean Dataset

Choose Principal Component Analysis

Compute the Dataset with the Reduced Attributes

End

As part of the PCA algorithm, the three components are orthogonal to each other. The orthogonal feature made the three components best vector to present the rest of the attributes which were not used in this work. The remaining seven attributes had a little correlation with each one of the three principal components.

Principal Component Analysis has thus mapped ten dimension dataset into a three dimension dataset. This dataset is used to train the Backpropagation Neural Network and Support Vector Machine to assess the effects of using PCA in attribute reduction applications of DGA in bushings fault detection.

### 2.1.5  Conclusion

In this section, the DGA bushings dataset of ten attributes is transformed into a low-dimension plane by using PCA. The experiment was conducted to choose the principal components, of which three components were chosen to represent the dataset.

The three principal components had acetylene, ethane and ethylene in the first component, carbon monoxide and total combustible gas in second principal component and nitrogen and oxygen gases attributes in the third component.

The accuracy of the chosen attributes is evaluated by assessing how the three attribute output affect the performance of the classifiers. To evaluate this, the BPNN and SVM classifiers are trained in the following chapter using the dataset obtained in this section.

## 2.2 Rough Set Theory

### 2.2.1 Background of Rough Sets

The theory was first introduced by Zdzislaw Pawlak [47] in the early 1980's. Rough Set theory has been applied in various applications within the machine learning domain and data mining domain. In most applications it was used to deal with missing data, making decision when the data is vague, dealing with high dimensional data, pattern identification and knowledge discovery. The work in [48-50] discussed the different applications where RS has been applied in data reduction applications as used in this work.

The section below discusses the mathematical background of the RS as it is applied in this work.

### 2.2.2 Mathematical Theory

This section discusses the mathematical formulas used in Degree of Dependency algorithm to reduce attributes of the DGA dataset. The algorithm was developed in five steps which are discussed mathematically below.

A. Information System

The DGA dataset is presented in a table with attributes in columns and bushings samples ID's in rows. This table is referred to as the Information System (IS), $\Lambda$ ,[51].

Let $U$ be a non-empty finite set of objects called the Universe.
Let $A$ be a non-empty finite set of attributes such that $a: U \to V_a$ for every $a \in A$.
The set of $V_a$ is called the value set of a
Then the Information System, $\Lambda$ is defined as

$$\Lambda = (U, A) \qquad \text{Equation (16)}$$

The DGA dataset in this work have an attribute called decision attribute. The dataset with a decision attribute in rough sets is called a decision system.

$$\Lambda = (U, A \cup \{d\})$$ Equation (17)

where $d \notin A$ is the decision attribute.

B. Data Discretization

The dataset to be analysed in Rough Sets need to be in categorical attribute values. Parametric method was used to discretise the attribute values into four categories. A categorical discretisation according to IEEE C57-104 [32] was used for the purpose of this experiment they are summarised in Table 2 below. The data was discretized as following the conditions defined Table 1 of the IEEE C57-104 standards.

Table 2: Categorical discretization according to IEEE C57-104 Standards as used in rough sets

| $H_2$ | $CH_4$ | $C_2H_2$ | $C_2H_4$ | $C_2H_6$ | $CO$ | $CO_2$ | $TDCG$ |
|---|---|---|---|---|---|---|---|
| 100 = 1 | 120 = 1 | 35 = 1 | 50 = 1 | 65 = 1 | 350 = 1 | 2500 = 1 | 720 = 1 |
| 101 – 700 = 2 | 121 – 400 = 2 | 36 – 50 = 2 | 51 – 100 = 2 | 66 – 100 = 2 | 351 – 570 = 2 | 2500 – 4000 = 2 | 721 – 1920 = 2 |
| 701 – 1800 = 3 | 401 – 1000 = 3 | 51- 80 = 3 | 101 – 200 = 3 | 101 – 200 = 3 | 571 – 1400 = 3 | 4001 – 10000 = 3 | 1921 – 4630 = 3 |
| >1800 = 4 | >1000 = 4 | >80 = 4 | >200 = 4 | >200 = 4 | >1400 = 4 | >10000 = 4 | >4630 = 4 |

Oxygen and Nitrogen are categorised by finding their normalising to a unit then $\{[0-.25]=1; [0.26-0.5]=2; [0.51-075]=3; [0.75-1]=4\}$ was used to categorise the oxygen and nitrogen attribute values.

C. Determine the Indiscernibility

Analysis using rough sets relies heavily on the indiscernibility of the attribute values for different observations.

**Definition 2.2**: A binary relation $R \subseteq XxX$ which is reflexive, symmetric and transitive is called an equivalence relation. Equivalence class of an element $x \in X$ consists of all objects $y \in X$ such that $xRy$

Let $\Lambda = (U, A)$ be the Information System, then with any $B \subseteq A$ there is an associated equivalence relation $IND_A(B)$ [52]

$$IND_A(B) = \{(x, x') \in U^2 | \forall a \in B \quad a(x) = a(x')\} \quad \text{Equation (18)}$$

$IND_A(B)$ is called the B-Indiscernibility Relation.

Equivalence class $[x]_B$ is a matrix with all undistinguished relations of B. The equivalence class used in this work included the decision attribute.

D. Approximations

Due to various factors in data acquisitions the values of the condition attributes for any two observations may be the same but with different decision attribute values. This problem can be experienced mostly after the data has been discretized.

To deal with uncertainties within the dataset, rough set theory introduces the approximations to group the data points of the dataset into two main groups. The two groups are lower approximations and upper approximations.

In the information system, if we let $B \subseteq A \; and \; X \subseteq U$ we can approximate $X$ using the information contained in $B$.

Lower Approximate

$$\underline{B}X = \{x| \; [x]_B \subseteq X\} \qquad \text{Equation (19)}$$

This group contains all observations which can be classified with certainty. These observations are called crisp sets.

Upper Approximate

$$\overline{B}X = \{x| \; [x]_B \cap X \neq 0\} \qquad \text{Equation (20)}$$

This group contains all observations which can just possibly belong to $X$. These observations are called rough sets.

Through the definitions of approximates, three regions within the dataset can be defined. The regions are boundary region, positive region and negative region. The regions are presented graphically in Figure 3.

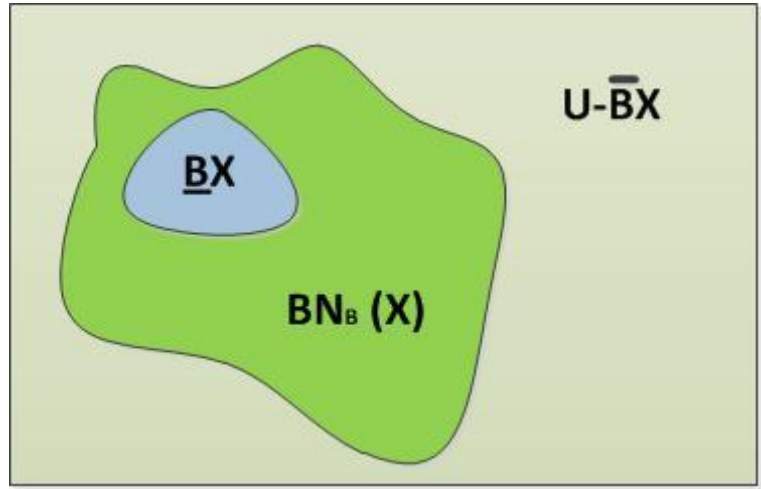
**Figure 3: Graphical presentation of the regions in Rough Set Theory**

i.  <u>Boundary Region</u>

This region is defined as the region between the upper approximates and the lower approximates of the dataset.

$$BN_B(X) = \overline{B}X - \underline{B}X \qquad \text{Equation (21)}$$

The observations in this region either belong to $X$ or they do not belong to $X$, thus they cannot be classified with certainty.

ii. <u>Positive Region</u>

This region is defined as the lower approximation region that contains all observations that can be classified with certainty.

$$B - Positive\ Region = \underline{B}X \qquad \text{Equation (22)}$$

   iii. <u>Negative Region</u>

This region is defined as the region that contains all observations not found in $X$.

$$B - Negative\ Region = U - \overline{B}X \qquad \text{Equation (23)}$$

The observations in this region do not belong to $X$ with certainty.

E. Degree of Dependency

**Definition 2.3**: Given Information System, $\Lambda$, a reduct of the information system is a minimal set of attributes $B \subseteq A$ such that $IND_\Lambda(B) = IND_\Lambda(A)$. The minimal set of attributes is called reducts.

Reducts have the same characteristics as the dataset that has all attributes. Let $D \subseteq A\ and\ C \subseteq A$. If the values of attributes from $C$ can be used to determine the unique values of attributes from $D$, then a set of attributes of $D$ depends totally on a set of attributes of $C$.

$$C \Rightarrow D \qquad \text{Equation (24)}$$

Thus there exists functional dependency between values of $D\ and\ C$ such that $D$ depends on $C$ in a degree of dependency $k$

$$C \Rightarrow_k D \qquad \text{Equation (25)}$$

Degree of dependency [21] is defined as a cardinality of the positive region over the cardinality of the universe.

$$k = \gamma(C, D) = \frac{\|POS_C D\|}{\|U\|} \qquad \text{Equation (26)}$$

where, $POS_C D$ is the positive region of partition $U/D$ with respect to $C$.

Degree of dependency range from $[0,1]$, such that if $k = 1$ then $D$ is totally dependent on $C$. If $k < 1$ then $D$ is partially dependent on $C$.

F. Attribute Reduction Using Degree of Dependency

Let $C \subseteq A, a \in C$, then $a$ is superfluous attribute if

$$U/IND(C) = U/IND(C - \{a\})$$  Equation (27)

The set $M$ is called minimal reducts of $C$ if and only if

i. $U/IND(M) = U/IND(C)$

ii. $U/IND(M) \neq U/IND(M - \{a\}), \forall a \in M$

If $C$ are conditional attributes and $D$ is a decision attribute, attributes $a \in C$ is called superfluous with respect to $D$ if

$$k = \frac{\gamma(C - \{a\}, D)}{\gamma(C, D)} = 1$$  Equation (28)

Otherwise $a$ is indispensable in $C$.

Thus subset $M$ of conditional attributes, $M \subseteq C$ must further satisfy the following conditions.

i. $k = \frac{\gamma(M, D)}{\gamma(C, D)} = 1$

ii. $k = \frac{\gamma(M-\{a\})}{\gamma(M,D)} \neq 1, \forall\, a\, \epsilon M$

### 2.2.3 Methodology and Experimental Setup

The aim of this experiment is to determine the minimum reducts using degree of dependency algorithm of the rough sets theory. The dataset in this experiment is prepared and discretized as shown in the mathematical equations in the above section. Figure 4 is a block diagram showing the general flow used to conduct this experiment.

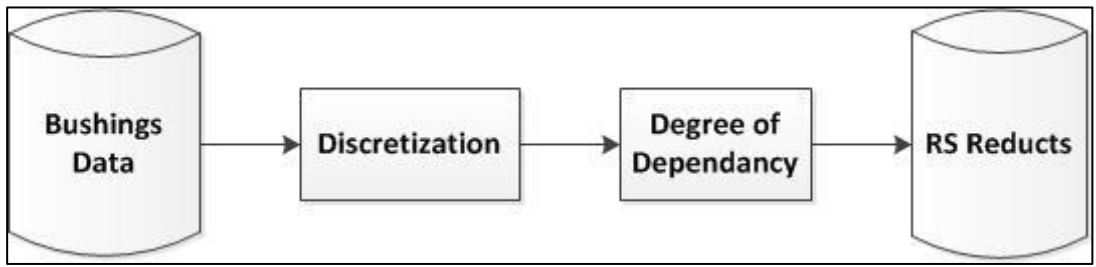

**Figure 4 : Block diagram illustrating the steps for Rough Set attribute reduction experiments**

The degree of dependency algorithm used in this work is shown below, Algorithm 2.2:

| Algorithm 2.2 - Degree of Dependency |
|---|

**Input:** DGA Dataset with 10 Conditional Attributes and 1 Decision Attribute

**Output:** Dataset with Reduced Attributes

**Begin:**
Discretise the Dataset

    **Function Degree of Dependency**

        Determine the Equivalence Class, $[x]_B$
$$IND_A(B) = \{(x, x') \in U^2 | \forall a \in B \quad a(x) = a(x')\}$$

        Find the Lower Approximation
$$\underline{B}X = \{x| \ [x]_B \subseteq X\}$$
        Find the Upper Approximate
$$\overline{B}X = \{x| \ [x]_B \cap X \neq 0\}$$
        Find positive Region
$$POS_C D = \underline{B}X$$
        Degree of Dependency
$$k = \gamma(C, D) = \frac{\|POS_C D\|}{\|U\|}$$
    **End**

    **Function Attribute Reduction**

        **For** N iteration until Number of Attributes-1
            Remove one Attribute
$$A - a$$
            Degree of Dependency
        **End**

        Compute Dataset of Attributes Combinations that have high a Degree of Dependency Value.
            Terminate Attribute Reduction when Degree of Dependency of full Attributes is Lower than Degree of Dependency of reduced Attributes Dataset.

    **End**

**End**

### 2.2.4 Results Discussion

The original dataset contained 10 Conditional Attributes and 1 Decision Attribute. The condition attributes were Acetylene, Carbon Dioxide, Carbon Monoxide, Ethane, Ethylene, Hydrogen, Methane, Nitrogen, Oxygen and Total Combustible Gases whose dependency on each other were analysed in this experiment.

The minimum attributes found to be indispensable in the gases-in-oil dataset were Acetylene, Carbon Monoxide, Ethane, Hydrogen, Methane and Total Combustible Gases. These results mean that these minimum gases should be enough to train the classifier and obtain accuracy higher or similar to the classifier trained using original dataset. The training time of the classifiers is also expected to decrease because few attributes are used to train the classifiers.

In the next chapter a dataset made of only the six attributes found to be indispensable by the degree of dependency is used to train classifiers and performance is compared as to how it affects the performance of SVM and BPNN. The performance is also evaluated when the SVM and BPNN performance trained using RS is compared to the performance of the same classifiers when using dataset reduced in other experiments in this chapter.

### 2.2.5 Conclusion

This section discussed the mathematical theory of rough sets' degree of dependency algorithm as it was employed in this experiment. In this section the experimental setup of the degree of dependency was also discussed. The results obtained shows that the degree of dependency reduced the 10 attribute conditions into six attributes.

## 2.3 Granular Computation

### 2.3.1 Background Theory of Granular Computation

Zadeh was the first to introduce the concept of Information granularity. He introduced it in the context of fuzzy sets where elements are considered within the granules than as individual entities [22]. Granular computation is thus a tool that deals with uncertain information, where in most engineering practices it is used during attribute reduction. Most researchers usually study and investigate granular computation based on a certain fuzzy technique; this could either be fuzzy logic or rough sets [53-55]. In this paper they are based on rough sets. The Incremental Granular Ranking (IGR or GR++) Algorithm is used in this paper.

### 2.3.2 Mathematical Theory

The incremental granular ranking algorithm used in this work was built on rough sets degree of dependency algorithm. To develop the incremental granular ranking algorithm, four steps are discussed mathematically.

### A. The Granular Ranking Algorithm

The data presented in this algorithm is prepared the same way the data is prepared for the rough sets analysis. If more information on how the data was prepared the reader can refer back to Section 2.2.

In summary five major properties of the rough sets are adopted to define the granular ranking algorithm [56-57]. These properties are Information System, Decision System, Positive Region, Negative Region and Boundary Region. The three regions are now called granules and are redefined below with notations that are used in the Granular Ranking Algorithm.

i. Positive Samples Granule

$$[x]_R^1 = \{y|\ x\epsilon\ POS_R(X), yRx\}$$  Equation (29)

ii. Boundary Samples Granule

$$[x]_R^2 = \{y|\ x\epsilon\ BN_R(X), yRx\}$$  Equation (30)

iii. Negative Samples Granule

$$[x]_R^3 = \{y|\ x\epsilon\ NEG_R(X), yRx\}$$  Equation (31)

## B. Rough Membership Function

Rough sets use the roughness membership function to determine the roughness of the granules. The rough membership universe [51] in rough sets theory is defined as follows:

$$\mu_X^R(x) = \frac{\|X \cap [x]_R\|}{\|[x]_R\|}$$  Equation (32)

The roughness in the dataset ranges from $[0,1]$, where the value differs for different granules. The roughness of different granules can be defined as follows.

i. Positive Samples Granule

$$\mu_X^R(x) = 1$$  Equation (33)

ii. Boundary Sample Granule

$$0 < \mu_X^R(x) < 1$$  Equation (34)

iii. Negative Sample Granule

$$\mu_X^R(x) = 0$$  Equation (35)

The granule sample with the highest number of positive samples is put in front of the other to optimize the results for instances where the rough membership of any two granules is the same.

### C. Granular Ranking Function

The granular ranking algorithm uses the granular ranking function that depends on the roughness of the dataset.

Let $\sigma_X^R(x) = \|(X \cap [x]_R)\|$ express the number of positive samples in the granule $[x]_R$. Then the Granular Ranking function, $r_G$ is defined as follows:

$$r_G(x) = \sigma_X^R(x) \times \mu_X^R(x) \qquad \text{Equation (36)}$$

$$= \|(X \cap [x]_R)\| \times \frac{\|(X \cap [x]_R)\|}{\|[x]_R\|}$$

$$= \frac{\|(X \cap [x]_R)\|^2}{\|[x]_R\|} \qquad \text{Equation (37)}$$

The granular ranking function value of any instance is also equal in instances where the rough membership function values are equal since the total number of positive samples, $\sigma_X^R(x)$, are also equal.

The granular ranking values are different for any regions. If $m = \|X\|$ then the granule region can be determined from the granular ranking function.

For instance $x$ in

i. Positive Sample Granule
   The granule ranking value is defined by the following range

$$1 \leq r_G(x) \leq m \qquad \text{Equation (38)}$$

ii. Negative Sample Granule

$$0 < r_G(x) < m \qquad \text{Equation (39)}$$

iii. Boundary Sample Granule

$$r_G(x) = 0 \qquad \text{Equation (40)}$$

D. **Incremental Granular Ranking Algorithm**

The Incremental Granular Ranking algorithm divides the dataset into smaller datasets. The granular ranking algorithm is then applied in the first small dataset, then after determining the granular ranking, the sample with the highest ranking function value is added as a new original dataset.

After the granular ranking value of the second dataset has been determined, the sample with the highest ranking function is added to the new original dataset. This is done until all smaller datasets have been ranked. To merge the two granules, the incremental granular ranking algorithm uses the combination algorithm. The combination algorithm is discussed in Section 2.3.3.

## 2.3.3 Methodology and Experimental Setup

The aim of this experiment is to use the degree of dependency algorithm through the incremental granular ranking algorithm to determine the minimum reducts that can be used to train the classifiers for fault detection.

The original dataset, like in other methods, has 10 condition attributes and one decision attribute. Figure 5 shows the block diagram that describes the stages followed in conducting this experiment.

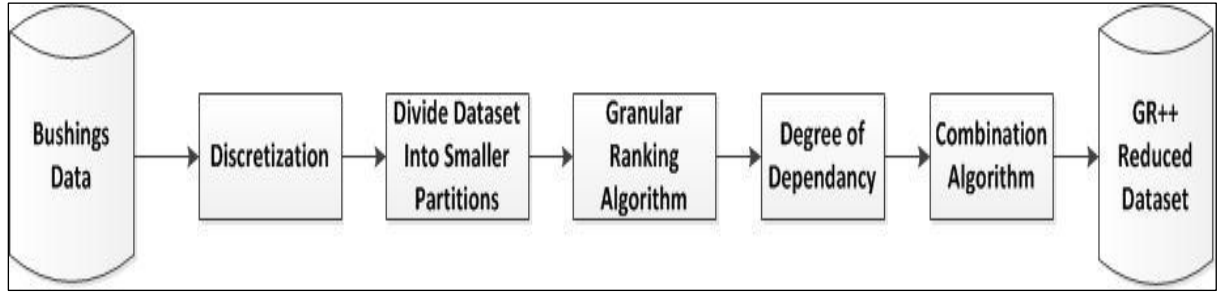

**Figure 5: Block Diagram Illustrating the Steps for Incremental Granular Ranking Attribute Reduction Experiments**

The degree of dependency algorithm used in this experiment is the same one as used in Section 2.2.3. The incremental Granular Ranking Algorithm as used in this experiment is indicated below, Algorithm 2.3.

## Algorithm 2.3 - Incremental Granular Ranking

**Input:** DGA Dataset with 10 Conditional Attributes and 1 Decision Attribute

**Output:** Dataset with Reduced Attributes

**Begin:**
Discretise the Dataset

    **Function Granular Ranking Algorithm**

        Find Positive Sample Granules
$$[x]_R^1 = \{y|\ x\epsilon\ POS_R(X), yRx\}$$

        Find Boundary Sample Granules
$$[x]_R^2 = \{y|\ x\epsilon\ BN_R(X), yRx\}$$

        Find Negative Sample Granules
$$[x]_R^3 = \{y|\ x\epsilon\ NEG_R(X), yRx\}$$

Find the Rough Membership of the Sample Investigated
$$\mu_X^R(x) = \frac{\|X \cap [x]_R\|}{\|[x]_R\|}$$

Apply the Granular Ranking Function
$$r_G(x) = \frac{\|(X \cap [x]_R)\|^2}{\|[x]_R\|}$$

Determine the Highly Ranked Sample

**End**

**Function Incremental Granular Ranking**

Input the Highly Ranked Sample

**Function Combination Algorithm**

**for** each Granularity instance $g'$ in the New Granular set $G'$
    **if** there is a correspondence granularity $g$ which has the same attribute values with $g'$, in the Granule Set $G$
    **then**
$$g.CountT = g.CountT + g'.CountT$$
$$g.CountF = g.CountF + g'.CountF$$
$$g.Proportion = \frac{g.CountT}{g.CountT + g.CountF}$$
$$g.r_G = g.CountT \times g.Propotion$$
**Else**
Insert $g'$ into $G$
**End**

Compute the Dataset with value from $G$
Use the Algorithm 2.2 to Reduce the Dataset Attributes of $G$.

Data of the new $g'$ shall have only attributes determined by Degree of Dependency.
**End**

**End**
**End**
**End**

### 2.3.4 Results Discussion

This algorithm reduced the number of attributes from ten attributes to four attributes. The four minimum attributes of the Incremental Granular Ranking algorithm experiment were Carbon Monoxide, Ethylene, Methane and Total Combustible Gases.

The dataset is then created from the original set where only these four gases are retained in the dataset that is then used to train the classifiers. The performance of the classifiers trained using GR++ minimum attributes dataset is compared to the performance of classifiers when other methods were used.

### 2.3.5 Conclusion

In this section the dimensions of the original dataset were reduced by use of the incremental granular ranking algorithm where four minimum reducts were obtained. The results of the G++ reducts differed with the reducts of the RS by two gases, viz. Acetylene and Hydrogen. The GR++ algorithm found this attributes to be superfluous attributes and the other four to be indispensable gases in the bushings oil.

## 2.4 Decision Trees

### 2.4.1 Theory of Decision Trees

Decision tree is a rule based classifier. There are different algorithms used in decision trees, viz. ID3, C4.5, CART, etc [66-69],
ID3 algorithm attempts to identify the attributes in the dataset that distinguishes classes. It uses the entropy function to measure how the attribute differentiates the classes.

$$\text{Entropy, } H(a) = \sum_A [-P(x_1|A)\log_2 P(x_1|A) - P(x_2|A)\log_2 P(x_2|A)] \qquad \text{Equation (41)}$$

$P(x_1|A)$ is the posteriori probability of $A$ in a population $x_i$

where $A = (a_1, a_2, a_3, \cdots a_n)$ n-dimensions with two classes.

Entropy is thus a measure of lack of order that exists in a system. When using conditional entropy, the most important features are the ones that give the lowest entropy. ID3 algorithm is however sensitive to gig numbers, as a result a new algorithm was introduced, the C4.5 algorithm [67]. The C4.5 algorithm is insensitive to big numbers and it uses information gain of which like entropy comes from the information theory.

Information gain, $I(X|Y)$ of a given attribute $X$ with respect to the class attribute $Y$ is the reduction in uncertainity about the value of $Y$ when we know the value of $X$. Entropy being used to measure the uncertainty about the value of Y.

$$I(Y|X) = H(Y) - H(Y|X) \qquad \text{Equation (42)}$$

This allows measuring the gain ratio:

$$\text{Gain Ratio } (Y|X) = \frac{I(Y|X)}{H(X)} \qquad \text{Equation (43)}$$

### 2.4.2 Methodology and Experimental Setup

The aim of this experiment is to use decision tree algorithm together with the pruning algorithms to determine the best attributes required to classify the DGA gases. The resultant decision tree was not used to classify unknown examples in this experiment. The MATLAB® decision tree algorithm was used in this work to find the best attributes.

The decision tree algorithm, Algorithm 2.4, in the MATLAB® Statistics toolbox is built on the C4.5 algorithm because it is capable of classifying both continuous and discrete data unlike the ID3 algorithm that can only classify the discrete data.

| Algorithm 2.4 - Decision Tree Algorithm |
|---|
| **Input:** DGA Dataset with 10 Conditional Attributes and 1 Decision Attribute<br><br>**Output:** Dataset with Reduces Attributes<br><br>**Begin:**<br>    Discretise the dataset<br><br>    Consider the entropy function $H(a_i)(i = 1, 2, \cdots n)$<br><br>    Calculate the entropy for each attribute<br>    Calculate the information gain<br>    Select the attribute with the highest information gain<br><br>    Divide the other attributes as sub-nodes for each selected attribute.<br><br>        **If** all objects in a sub-node belong to the same class<br>        Then that sub-node is a terminal node<br><br>        **Else**<br><br>        Find the entropy of the various attributes in the sub-node<br>        The attribute with the highest information gain is thus the sub-node of that point.<br><br>        Repeat this algorithm until all sub-nodes are terminal nodes.<br><br>        **End**<br>**End** |

### 2.4.3 Result Discussion

The experiment conducted in this section resulted in three attributes being chosen by the decision trees. These are carbon dioxide, ethane and total combustible gases. The dataset is then created using these three gas attributes.

The created dataset is used to train the Backpropagation neural network and the support vector machine classifier.

## 2.5 Conclusion

This chapter focused on methods used to reduce attributes in the DGA dataset. The attributes were reduced by examining the correlation between gas attributes and dependency relationship between gas attributes and decision attribute. The principal components analysis was used to find the correlation between gases, where it revealed that the first principal had acetylene, ethane and ethylene high correlated and with the greatest variance. The carbon monoxide gas attribute was correlated to the combination of combustible gases attribute.

The dependency relation was examined using Rough Set Theory, Incremental Granular Ranking and Decision Tree algorithms. These algorithms determined ethane and total combustible gas attributes to be the core attributes in determining the dependency between gases and decision. The table below shows a summary of gases identified as core attributes by each algorithm.

Table 3: Table summarising the gas attributes chosen by the different preprocessors as the core gases in decision making

|      | $C_2H_6$ | $CO_2$ | $CO$ | $C_2H_2$ | $C_2H_4$ | $H_2$ | $CH_4$ | $N_2$ | $O_2$ | TCG |
|------|----------|--------|------|----------|----------|-------|--------|-------|-------|-----|
| RS   | X        |        | X    | X        |          | X     | X      |       |       | X   |
| GR++ |          |        | X    | X        |          |       | X      |       |       | X   |
| PCA  | X        |        | XX   | X        | X        |       |        | XXX   | XXX   | XX  |
| DT   |          | X      |      | X        |          |       |        |       |       | X   |

In the above table 'X' indicates the gases identified as core gases by each algorithm. In the PCA algorithm, the gases that belong to the first principal are indicated by 'X', the second principal component gasses are 'XX' with the 'XXX' indicating the ones that belong to the third component.

These experiments were used on gases of two class problem that only indicated whether the transformer was faulty or non-faulty. According to IEEE standards, hydrogen gas is formed in most of the faults conditions. Most of these algorithms didn't choose hydrogen as a distinguishing attribute that can be used to make decision because there are traces of it in almost all conditions. This finding supports a study done by [68-69] that addressed the misconceptions that people make when it comes to DGA. The misconceptions are that a single gas such as acetylene can be used to diagnose arc related faults, carbon monoxide can be used to diagnose the degrading of cellulose and hydrogen gas is important since it is found in all fault conditions. In the study found in [69], the authors showed that these assumptions are wrong since they do not reveal the true nature of the arc and the presence of other gases can be used to reveal it.

The study also showed that the fact that hydrogen is present in most conditions disqualifies it as a diagnostic gas. It was also revealed that even when used in ratio with other gases, hydrogen gas was not as informative as methane gas. The other reason why hydrogen gas was least not chosen the most by these preprocessors is that hydrogen gases are least soluble in oil with high diffusion rate. These thus lower its reliability in diagnostic purposes since its concentration levels in samples lower at faster rates. Ethane was however chosen as the most preferred gas on which the decision attribute can depend on in this work by the preprocessors.

The outputs of each method are then used in the following chapter to assess the effects of each method; in transformer bushings DGA applications. There are four computational intelligent method investigated for this application; namely principal component analysis, rough set, incremental granular ranking and decision tree. The background theory of each method was discussed, followed by the way in which they were setup for experiments in this works.

The results obtained from each experiment were discussed followed by the conclusion statement for each experiment. The output results of each method were not compared in this chapter as this is done in the following chapter following their effects in classifiers performances.

# 3 CLASSIFIERS FOR DISSOLVED GASES-IN-OIL ANALYSIS DATASET

## 3.1 Backpropagation Neural Networks

### 3.1.1 Background Theory of Neural Networks

An artificial neural network is a mathematical model that mimics the human nervous system. Neural networks have been adapted to many different fields since their introduction. These fields include pattern recognition, data analysis, speech recognition, data mining, etc. in this paper they are adapted to data analysis.

Neural Networks are very robust to noise [13] unlike rough sets, but not effective in processing high dimensional data sets. The accuracy of the neural classifications tends to be very low when they process high dimensional data for classification. Applications of artificial neural networks in different condition monitoring applications can be seen in [11-23, 60-62]

### 3.1.2 Mathematical Theory

The total number of condition attributes determines the number of input neurons in a neural network structure. The input dataset in neural networks is normalised so that other attributes are not weighed more than the other attributes because of the range of their values. The output of neural networks is a subject to the input data, activation function and weights.

$$f = g(x, w) \qquad \text{Equation (44)}$$

where $g$ is the activation function, $x$ is the input and $w$ is the weight.

The activation functions are used to determine the output of the neural network. The log-sigmoid and tan-sigmoid activation functions are used in this work.

A. Log-sigmoid Function

$$g = \frac{1}{1 - e^{-n}}$$ Equation (45)

The output of the log-sigmoid function range from $[0,1]$

B. Tan-Sigmoid Function

$$g = \frac{e^n - e^{-n}}{e^n + e^{-n}}$$ Equation (46)

The output of the tan-sigmoid function ranges from $[-1,1]$

The log-sigmoid function was used to calculate the output of the output layer whereas the tan-sigmoid function was used to calculate the output of hidden layers.

The Backpropagation algorithm was used to train the neural networks in this work. The neural networks are trained by calculating the output error and using it to update the weights.

The error is the represented as a sum of square of the output and target difference.

$$Error = \sum_j (output - target)^2$$ Equation (47)

$$= \sum_j (O_j - d_j)^2$$

The error is bigger if the difference between the output and the target is large and lesser if the difference is small. In Backpropagation, the dependency of the error on input, output and weights is determined by the gradient descendent method.

$$\Delta w_{ji} = -\eta \frac{\partial E}{\partial w_{ji}} \qquad \text{Equation (48)}$$

$$= -2\eta (O_j - d_j) O_j (1 - O_j) x_i$$

$\eta$ is a constant called a learning rate.
$\Delta w_{ji}$ is the value used to adjust weights.

Backpropagation algorithm uses the $\Delta w_{ji}$ to update the weights and retrain the neural networks. Since calculating the error in the hidden layer cannot be calculated directly because it does not have a decision, the error from the output layer is propagated back to the hidden layer.

$$Error_{HiddenLayer} = \sum_j \left[ (O_j - d_j)^2 (Error_{OutputLayer} \times w_{ji}) \right] \qquad \text{Equation (49)}$$

To update the hidden layer weights, the following function is used:

$$(UpdatedWeights)_{Hidden-Output} = InitialWeight + Error_{HiddenLayer} \times Input \qquad \text{Equation (50)}$$

The output is then calculated using the updated weights. This process is repeated until the desired minimum error is achieved or once the validation set's error is at a minimum.

The desired minimum error is a method used to stop training the neural network. This is achieved by adding the absolute values of the errors from each neuron. The other better method is partitioning the training dataset into a separate set of validation sets. The validation set is used to calculate the error of classifications every time after the network has been trained with updated weights.

When using the validation set, the neural network stops training once the validation set error starts rising. In this work to optimize the error and reduce the local minima cases, the validation set error was set to stop training the network if the error increased for six consecutive iterations.

### 3.1.3 Methodology and Experimental Setup

The experiments were done in four parts where the BPNN was trained first using raw unpreprocessed dataset. During this training the training parameters that result in the highest classification accuracy was saved and used as configuration parameters for the rest of the experiment. These experiments were performed in the MATLAB® platform with the Neural Network Toolbox.

To maximize the number of iterations during training 1000 iterations were used to propagate the error until the desired minimum error of $e^{-5}$ was obtained. The neural network was designed to stop training when the desired error is obtained. In real applications, the neural network training might not obtain the minimum error; in such cases the neural network over-fits or under-fits. This work used the validation set method to avoid the over-fitting and under-fitting problems.

The training set was split into a validation set; the validation set is used to train the BPNN model at the end of iteration. If the output error starts increasing for six consecutive iterations, the neural network stops the training process. The validation set method however has a validation of which is not addressed in the scope of this work. This drawback is that there is no way to identify whether the iteration at which the training stops at is a local or global minima of the classification error.

Cross-Validation was used to reduce the biasness of the dataset during the training phase. This work used 15 folds in the cross-validation.

Figure 6 shows an overview of the method used to evaluate the effects of preprocessing the data before training the BPNN with the bushings fault dataset.

The BPNN is trained first using the unpreprocessed data which has 10 conditional attribute and bi-class fault decision. Therefore the neural network was designed to have 10 inputs. During the design of the BPNN architecture it was found that the neural network performed better using two hidden layers. The neurons of the second hidden layer improved the BPNN performance significantly when they are more than the neurons of the first hidden layer.

BPNN performance improved when the tan-sigmoid activation function was used during the experiment. The log-sigmoid function was used in the output layer because the original decisions range from [0,1]. The final configuration obtained from training the unpreprocessed data was then used to train the BPNN using the preprocessed dataset.

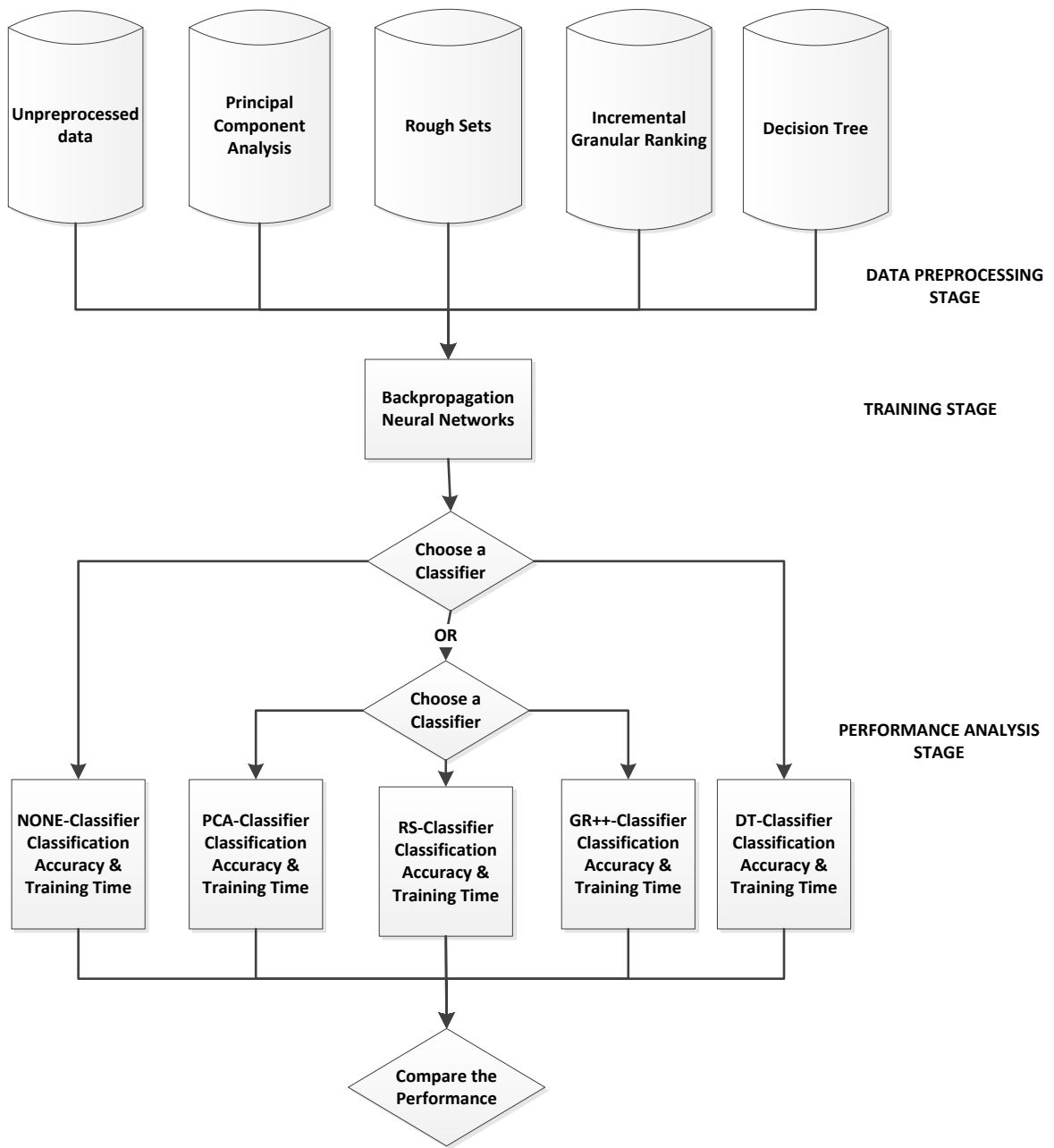

**Figure 6: flow diagram showing how the dataset from preprocessing techniques was used to train the classifiers**

The final configurations used to train the BPNN in these experiments are shown in the Table 4 below.

**Table 4 : The configuration parameters used to train the Backpropagation Neural Networks**

| Parameters | Configuration |
|---|---|
| Epochs | 1000 |
| Learning Rate | 0.05 |
| k-folds | 15 |
| Training Data Ratio | 0.7 |
| Test Data Ratio | 0.15 |
| Validation Data Ratio | 0.15 |
| Training Goal | $e^{-5}$ |
| Training Function | traingd |
| Number of Hidden Layers | 2 |
| Neurons in Hidden Layer 1 | 20 |
| Neurons in Hidden Layer 2 | 30 |

The BPNN performance was evaluated in terms of accuracy and training time. The results obtained when the configurations in Table 4 above were used to train unpreprocessed data is shown in Table 5 below.

**Table 5: Experimental results of BPNN classifier trained using unpreprocessed dataset**

| NONE-BPNN | | |
|---|---|---|
| k-Folds | Average Accuracy (%) | Average Training Time(s) |
| 15 | 91.6 | 59.93 |

The training parameter configurations were kept constant throughout the experiments so that the effects of each preprocessing method can be evaluated on the same common ground.

The dataset discussed as the output of PCA in Section 2.1 was used to train the BPNN. This classifier was then called PCA-BPNN whose performance results are shown in Table 6.

**Table 6: Experimental results of BPNN classifier trained using principal component analysis preprocessed dataset**

| PCA-BPNN | | |
|---|---|---|
| k-Folds | Average Accuracy (%) | Average Training Time(s) |
| 15 | 96.7 | 49.21 |

The other part of the experiment was to test the performance of the BPNN when using RS dataset as discussed in Section 2.2. The training performance of the RS-BPNN is indicated in Table 7.

**Table 7: Experimental results of BPNN classifier trained using rough set preprocessed dataset**

| RS-BPNN | | |
|---|---|---|
| k-Folds | Average Accuracy (%) | Average Training Time(s) |
| 15 | 86.1 | 27.84 |

The next part of the experiment in evaluating the effects of preprocessing for BPNN was training the BPNN classifier using the dataset obtained as the output of the GR++ as discussed in Section 2.3. Table 8 below shows the results obtained during this part.

**Table 8: Experimental results of BPNN classifier trained using incremental granular ranking preprocessed dataset**

| GR++-BPNN | | |
|---|---|---|
| k-Folds | Average Accuracy (%) | Average Training Time(s) |
| 15 | 89.5 | 18.28 |

The last part of the experiment was to train the BPNN classifier using the data obtained when preprocessed using decision trees. The results obtained in the experiment are discussed below in Table 9

Table 9: Experimental results of BPNN classifier trained using decision trees preprocessed dataset

| DT-BPNN | | |
|---|---|---|
| k-Folds | Average Accuracy (%) | Average Training Time(s) |
| 15 | 92.4 | 12.47 |

### 3.1.4 Results and Discussion

The performance of the classifiers is evaluated based on the classification accuracy. Figure 7 below is a graph showing the accuracy performance of the BPNN when different datasets from the investigated preprocessors is used.

The BPNN trained using the dataset from the PCA preprocessor shows an improvement in classification accuracy whereas the classification accuracy when the RS dataset has the lowest classification accuracy.

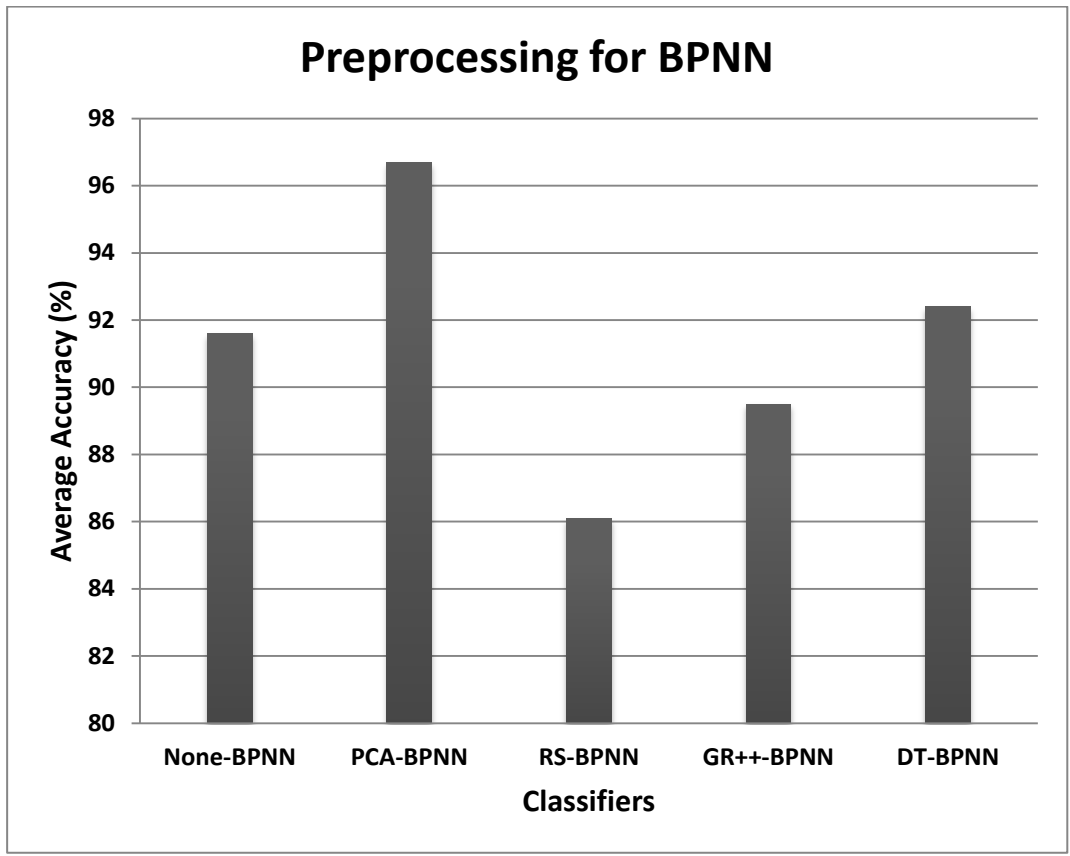

**Figure 7 : Bar graph of classification accuracy of BPNN classifiers when trained with different dataset**

The training time graph for the BPNN classifier is shown in Figure 8 below. The first bar in Figure 8 shows the training time of BPNN when dataset with full ten attributes was used for training. The training time decreased when the dataset with reduced attributes where used for training.

However performance of BPNN under different preprocessors cannot be evaluated by just looking at the accuracy or training time separately. For a better performance measure, the balance of the two was used. In this experiment it shows that the BPNN improves in both accuracy and training time when the dataset from PCA preprocessor is used for training.

Even though the GR++ dataset decreases the training time, it is at the expense of the accuracy.

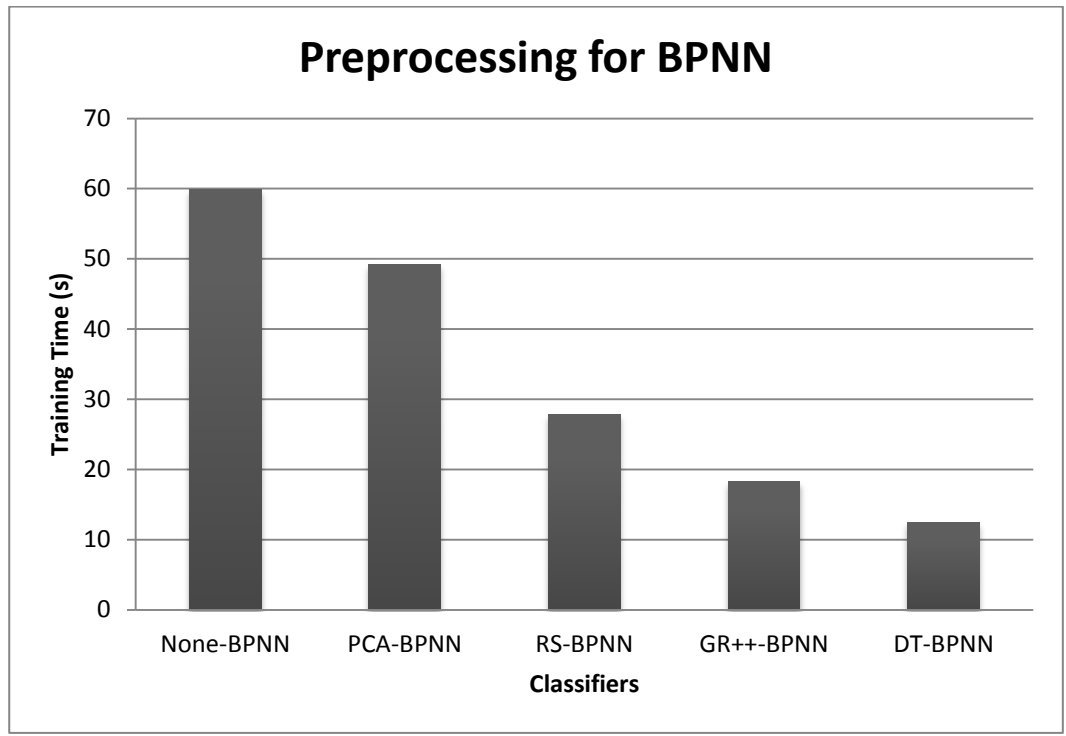

**Figure 8: Bar graph of average training time of BPNN classifiers when trained with different dataset**

### 3.1.5 Conclusion

In this experiment the BPNN is evaluated when trained using dataset from different preprocessors for the bushings fault data. The graphs clearly indicate that the use of Principal Component Analysis improved the performance of the BPNN classifier in terms of accuracy and training time. From this experiment it can be concluded that if the BPNN is to be used for classification of bushings DGA dataset, PCA must be used to reduce the attributes of the original dataset.

The results obtained here are compared to the results obtained from the next experiment where the best preprocessor-classifier combination can be deduced. The next experiment uses the same dataset used to train the BPNN to train the SVM classifiers.

## 3.2 Support Vector Machines

### 3.2.1 Background Theory of Support Vector Machines

Support Vector Machine is a statistically based machine learning technique which was introduced by Cortes and Vapnik in 1995. Vapnik then modified it in 1999. SVM classifies data by creating a hyperplane that divides the data that belongs to one class from that that don't belong.

The aim of the SVM is increasing the margin between the classes, it does this by creating a hyperplane and the data points closer to these planes are called support vectors. By finding the optimal hyperplane that creates a large margin between the classes, it minimizes the generalisation error. The SVM has been applied by different researchers for condition monitoring [9, 23-30, 64-65]

### 3.2.2 Mathematical Theory

With the assumption that the data is linear, if we have two dimensions we can use a straight line to separate the two dimensions data of two classes. In cases of higher dimensions, we use hyperplanes to separate the data [24-26]. Support Vectors are the data points that lie closest to the hyperplane.

Support Vector Machines separate the data points into two different classes according to the margins defined as

$$(w.x) + b = 1 \; for \; hyperplane \; 1 \qquad \text{Equation (51)}$$

$$(w.x) + b = -1 \; for \; hyperplane \; 2 \qquad \text{Equation (52)}$$

The SVM then places a boundary line between the marginal in such a way that the support vectors are the maximally away from the boundary line.

Support Vectors lying on the marginal contain all necessary information needed for classification of the data. The boundary line is defined

$$(w.x) + b = 0 \quad w \in R^D, b \in R \quad \text{Equation (53)}$$

Vector $w$ defines the boundary
$x$ is the input vector of dimension $D$

Using the Support Vectors defined as points that lie on the margins, the classification decision function can be defined as

$$f(x) = sign\big((w.x) + b\big) \quad \text{Equation (54)}$$

The optimal hyperplane is tested by satisfaction of the following functions

$$\tau(w) = \frac{1}{2}\|w\|^2 \quad \text{Equation (55)}$$

$$y_i\big((w.x_i) + b\big) \geq 1; \quad i = 1 \cdots l \quad \text{Equation (56)}$$

where $l$ is the number of training sets.

Transformation is done in non-linear cases where the above equations are not satisfied by the hyperplane. In such cases the data is mapped into a higher dimension where a linear separation that can satisfy the above equations is found.

Let $w = \sum v_i x_i$
where $v_i$ represents the weights

$$\therefore f(x) = sign\left(\sum v_i(x.x_i) + b\right) \quad \text{Equation (57)}$$

Input space of D-Dimensions is transformed into feature space of Q-Dimensions.

$$s = \emptyset(x) \quad x \in R^D \ \& \ s \in R^Q \qquad \text{Equation (58)}$$

$$\therefore f(x) = sign\left(\sum v_i(\emptyset(x) \cdot \emptyset(x_i)) + b\right) \qquad \text{Equation (59)}$$

The transformation of input space to higher dimensional space and the dot product can be performed inside a kernel to improve the computational performance.

A kernel can be defined as

$$k(x,y) = \emptyset(x) \cdot \emptyset(y) \qquad \text{Equation (60)}$$

$$\therefore f(x) = sign\left(\sum \left(v_i(k(x, x_i))\right) + b\right) \qquad \text{Equation (61)}$$

There are different kernels for SVM, viz. polynomial, sigmoid and radial basis function kernels.

### 3.2.3 Methodology and Experimental Setup

SVM was used to assess the effects of preprocessing the bushings dataset by reducing the attributes. The SVM were used in this work in a similar way the BPNN classifier was used. SVM trained first using unpreprocessed data and the configuration that produces the best results is used to train using dataset from different preprocessors. The experiments were conducted in the MATLAB® platform with the Bioinformatics Toolbox.

Cross-Validation of 8-folds was used to reduce the biasness during the training of the SVM classifier during the experiments.

Average classification accuracy and average training time was noted when training SVM classifier using the raw unpreprocessed data from the bushings. The obtained results are recorded in Table 10 below.

Table 10: Experimental results of SVM classifier trained using unpreprocessed dataset

| None-SVM | | |
|---|---|---|
| k-Folds | Average Accuracy (%) | Average Training Time(s) |
| 8 | 87.8 | 25.56 |

After the best SVM classifier configuration was obtained, the classifier was then trained using the dataset obtained from the PCA. The description of the dataset obtained from PCA is described in Section 2.1. The classifier's performance results are shown in Table 11 below.

Table 11: Experimental results of SVM classifier trained using principal component analysis preprocessed dataset

| PCA-SVM | | |
|---|---|---|
| k-Folds | Average Accuracy (%) | Average Training Time(s) |
| 8 | 93.5 | 8.62 |

The classifier was then trained using the dataset obtained from the Rough Set described in Section 2.2. The performance results are in Table 12 below.

**Table 12: Experimental results of SVM classifier trained using rough sets preprocessed dataset**

| RS-SVM | | |
|---|---|---|
| k-Folds | Average Accuracy (%) | Average Training Time(s) |
| 8 | 86.9 | 27.34 |

The next experiment was training the SVM classifier of this experiment with the dataset obtained from the GR++ preprocessor. The results are displayed in Table 13.

**Table 13: Experimental results of SVM classifier trained using incremental granular ranking preprocessed dataset**

| GR++-SVM | | |
|---|---|---|
| k-Folds | Average Accuracy (%) | Average Training Time(s) |
| 8 | 85.2 | 14.23 |

The final experiment was to test the performance of the SVM when trained using the dataset obtained by preprocessing the DGA data using decision trees.

**Table 14: Experimental results of SVM classifier trained using decision tree preprocessed dataset**

| DT-SVM | | |
|---|---|---|
| k-Folds | Average Accuracy (%) | Average Training Time(s) |
| 8 | 89.0 | 8.17 |

### 3.2.4 Results and Discussion

The graph in Figure 9 displayed below shows the classification accuracy of the SVM classifier when trained using dataset from different preprocessors.

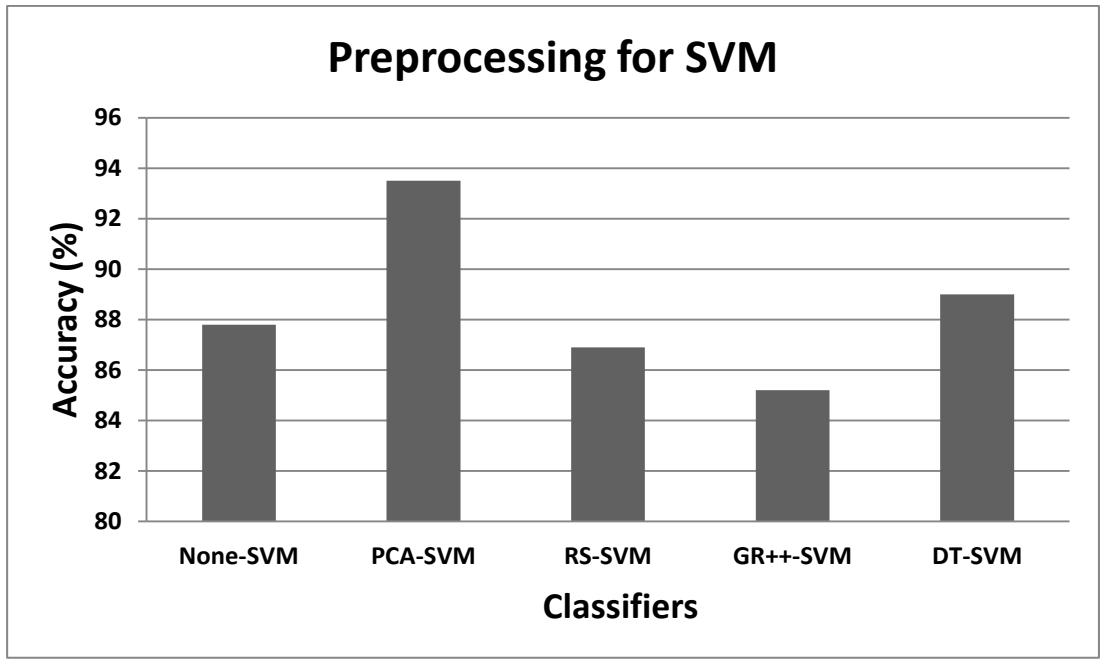

**Figure 9: Bar graph of classification accuracy of SVM classifiers when trained with different dataset**

Principal Component Analysis preprocessor improved the classification accuracy of the SVM classifier. When trained with the dataset obtained through the rough set based methods, the SVM classifier's classification accuracy decreased.

The evaluation based on training time is indicated in the graph displayed in Figure 10 below for SVM classifiers.

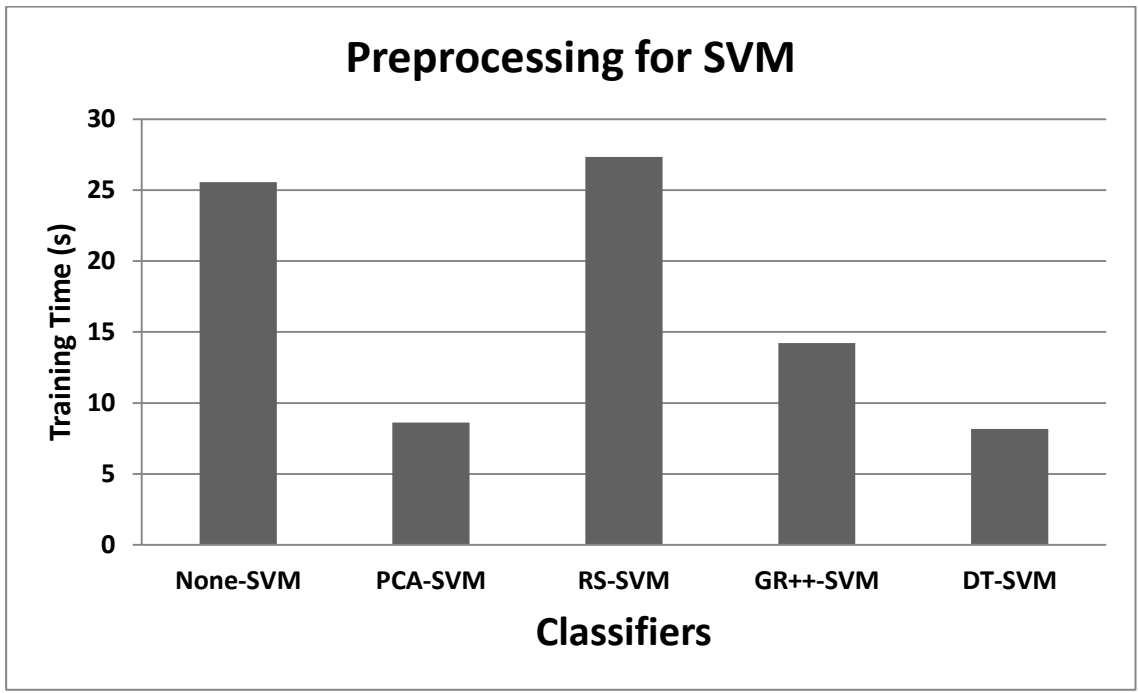

**Figure 10: Bar graph of average training time of SVM classifiers when trained with different dataset**

Training the SVM classifier with the dataset obtained from the RS preprocessor increased the training time and decreased the classification accuracy, whereas the PCA dataset increased the accuracy and decreased the training time.

### 3.2.5 Conclusion

The use of PCA to preprocess the bushings DGA dataset improved the performance of the SVM classifier. Rough sets preprocessor for bushings DGA performed poorly as they worsen the SVM performance than the dataset from the GR++ preprocessor. When compared with the results of the BPNN, PCA preprocessors produces a dataset that improves both the training time and accuracy of the classifiers whereas the Rough Sets based preprocessors performed poor than the unpreprocessed dataset.

PCA thus makes the best bushings DGA preprocessors when using BPNN and SVM classifiers for fault detection in bushings. GR++ preprocessor to reduce the attributes for the bushings only helps decrease the time the classifiers used to learn the patterns but compromised their classification accuracy with a big margin. It thus wouldn't be advisory to use GR++ in reducing attributes of bushings DGA dataset.

## 3.3 Chapter Conclusion

The experiments conducted in this chapter indicate that the classification and learning performance of classifiers in bushings fault detection can be improved by using preprocessors to reduce the number of attributes. The different attribute reduction methods used reduced different attributes where the PCA works more as a transformation technique where multi-dimensional dataset is transformed into lower-dimensions. The different experiments conducted in Chapter 2 showed which attributes were the core attributes. Those core attributes were used to train the BPNN and SVM classifiers.

The PCA-BPNN combination triumphs the PCA-SVM combination because the former has a very high accuracy. The latter, however learns faster. In this work the classifier that learns faster but classifies poorly was considered as a poor classifier. Training speed on its own with poor classification accuracy is poor because the classifier becomes bad in generalising new unseen data. Thus the GR++-BPNN and GR++-SVM combination are the poorest combination because they only learn fast and classify poorly. Thus the GR++ was the poorest attribute reduction in this application.

The RS-BPNN and RS-SVM combinations were considered as poor combinations because they reduced the classification accuracy of the classifiers. This means that the RS preprocessor was also a poor preprocessor in this application since the classifiers' classification accuracy dropped after learning using the RS dataset.

# 4 INTRODUCTION OF ROUGH NEURAL NETWORKS AS CLASSIFIERS IN BUSHINGS FAULTS ANALYSIS USING DISSOLVED GAS-IN-OIL ANALYSIS

## 4.1 Theory of Rough Neural Networks

Rough Neural Network was introduced by Lingras in 1996 in the traffic volume prediction application. Rough Neural Network is a hybrid of Rough Set and Neural Network. RS partitions the dataset into different regions which are subsets of the universe. A neuron of the Neural Network is trained using the data that belongs to each one of the regions. A neuron trained using data in the regions is then called a rough neuron.

The rough neural network uses the upper approximates and lower approximates of the rough sets to construct an upper rough neuron $\overline{r}$ and a lower rough neuron $\underline{r}$. Rough Neural Networks thus exhibit the good properties of the rough set and the neural network model [36-42, 58-61].

The upper neuron $\overline{r}$ is trained with the dataset of the upper approximate which contain all observations that might be contained in a dataset, i.e. rough sets.
The output of the upper rough neuron is determined as follows:

$$O_{\overline{r}} = \max\left(g(\overline{x}), g(\underline{x})\right) \qquad \text{Equation (62)}$$

The lower neuron $\underline{r}$ is trained with the dataset of the lower approximates which contains all observations that certainly belong in a dataset, i.e. crisp dataset.
The output of the lower rough neuron is determined as follows:

$$O_{\underline{r}} = min\left(g(\overline{x}), g(\underline{x})\right) \qquad \text{Equation (63)}$$

Two rough neurons can be connected in three possible ways using either four or two connections [36]. The three different connections are:

1. Full Connection

    Fully connected rough neurons have four connection points.

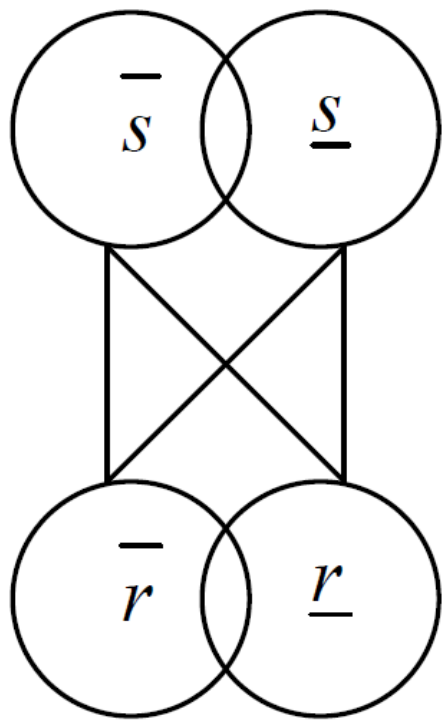

**Figure 11: Fully Connected Rough Neurons with four connections**

2. Excitatory Connection

    There are only two connections in the excitatory connection of rough neurons where the increase in the output of rough neuron $r$ results in the increase in the output of rough neuron $s$. Thus rough neuron $r$ is said to excite activities in rough neuron $s$. The excitatory connection is shown in Figure 12.

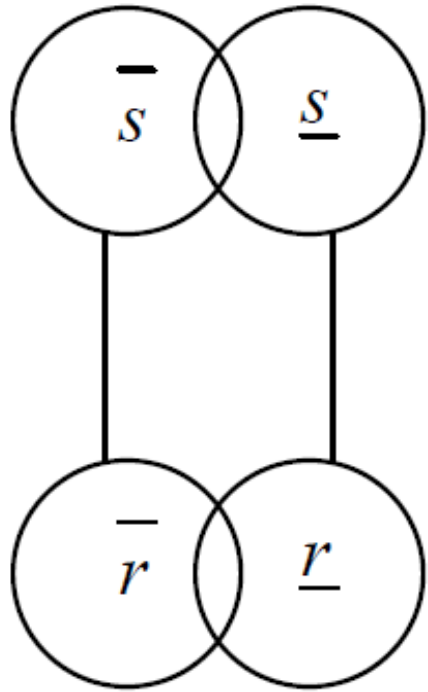

**Figure 12: Excitatory connection of two rough neurons with two connections**

3. Inhibitory Connection

   Inhibitory connected rough neurons also have just two connections. Connection is shown in Figure 13. The increase in the output of rough neuron $r$ corresponds to the decrease in the output of rough neuron $s$. Thus rough neuron $r$ is said to inhibit the activities in rough neuron $s$.

Rough neurons are connected to the conventional neurons using only two connections. The excitatory connection is used to connect the input rough neurons to the hidden layer rough neurons in this work.

It should be noted that it is physically impossible to connect the upper and lower rough neurons hence their simulations are done separately but the output is added.

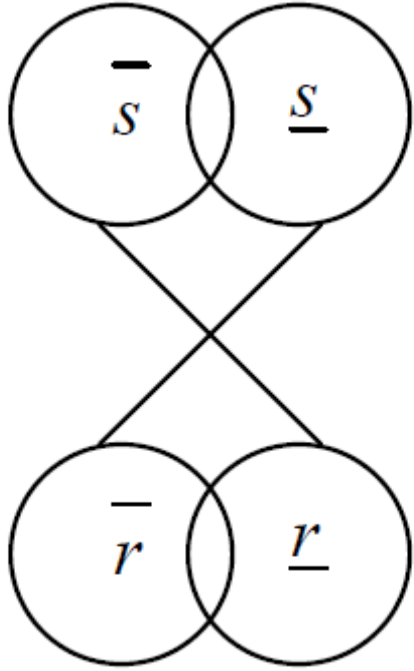

**Figure 13: Inhibitory connection of two rough neurons with two connections**

The output neuron used is the conventional neuron with summation capabilities. The final output is thus

$$Output = O_{\bar{r}} + O_{\underline{r}}  \quad \text{Equation (64)}$$

To present the output with some summation function $f$.

$$Output = f(O_{\bar{r}} + O_{\underline{r}}) \quad \text{Equation (65)}$$

Sandeep et al. proposed using boundary region and lower approximation subsets to train a rough neuron, thus called New Rough Neural Network [52]. The output of the boundary rough neuron is computed using the following function:

$$O_{\bar{r}-\underline{r}} = \frac{g(\underline{x})}{\varepsilon_{\underline{r}}} + \frac{g(x_{\bar{r}-\underline{r}})}{\varepsilon_{\bar{r}-\underline{r}}} \quad \text{Equation (66)}$$

Where $\varepsilon_{\bar{r}-\underline{r}}$ is the Output Excitation Factor for Boundary Neuron calculated using the following equation:

$$\varepsilon_{\bar{r}-\underline{r}} = -\left[\frac{\Delta(O_{m_i})}{\Delta(x_{m_i})}\right] \qquad \text{Equation (67)}$$

And $\varepsilon_{\underline{r}}$ is the Output Excitation Factor for Lower Neuron calculated using the following equation:

$$\varepsilon_{\underline{r}=}\left[\frac{\Delta(O_{m_i})}{\Delta(x_{m_i})}\right] \qquad \text{Equation (68)}$$

In the above equations:
- $\Delta(x)$ is the change in input
- $\Delta(O)$ is the change in output
- $m$ is the $mth$ layer in neural network
- $i$ is the $ith$ node in $m$.

The RNN inherits the ANN architecture in that it consists of three layers, viz. input layer, hidden layer and output layers. The rough neurons can be used in the input layer only, as shown in Figure 14:

In other applications the rough neurons can be implemented in the hidden layer too. The architecture of an RNN with rough neurons in hidden layer is shown in Figure 15.

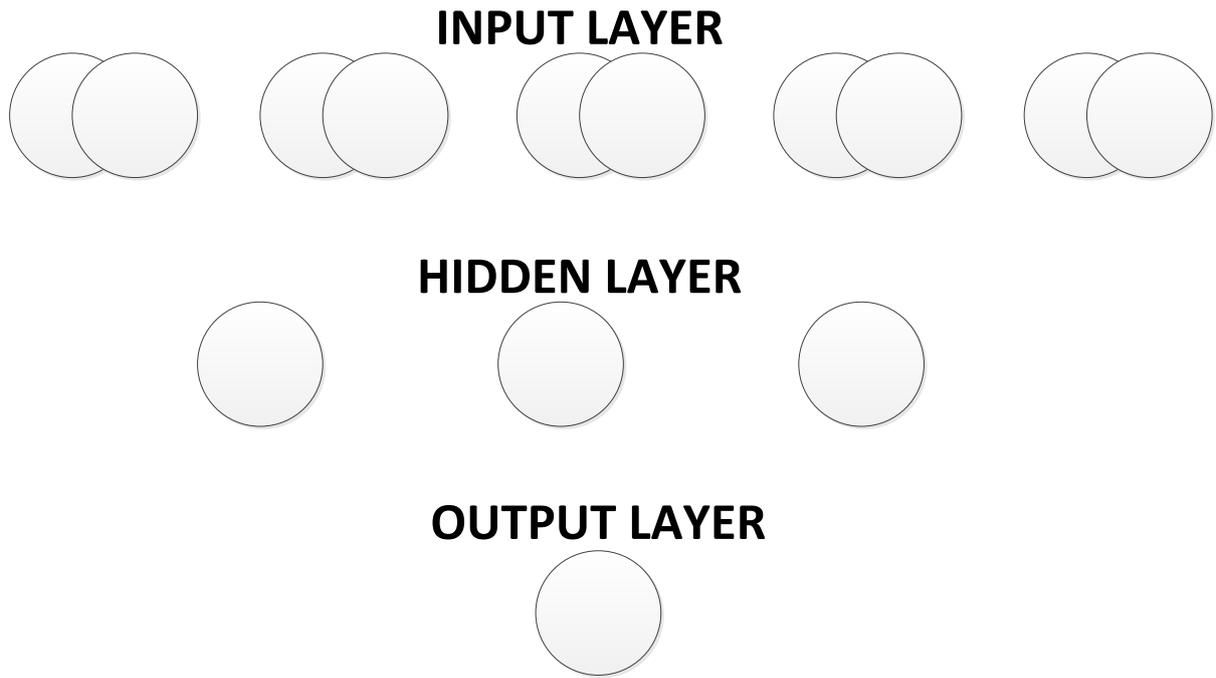

Figure 14: Rough Neural Network architecture with rough neurons in the input layer only

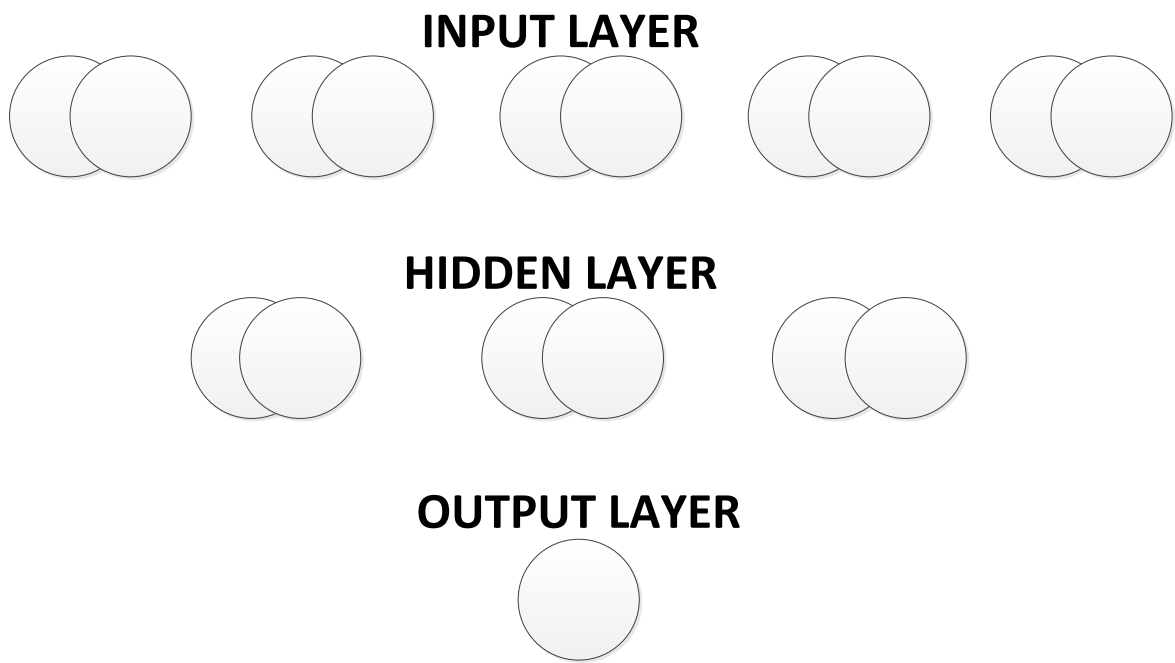

Figure 15: Rough Neural Network architecture with rough neurons in the input layer and the hidden layer

The Rough Neural Network implemented in this work uses the Lower and Upper Rough Neurons only. The Backpropagation Neural Network was used to develop the rough neural networks in this work using the architecture showed in Figure 14 where the rough neurons are connected using the excitatory connection.

## 4.2 Methodology and Experimental Setup

### 4.2.1 Training RNN Using Unpreprocessed Dataset

To assess the classifier's performance, it was trained using raw unpreprocessed data where the performance results were compared to the performance of the BPNN and SVM which were trained in the experiments conducted in Section 3.1 and Section 3.2 respectively using the same unpreprocessed dataset.

The methodology followed in this part of the experiment where the RNN performance is assessed against the BPNN and SVM performance is similar to the one shown in Figure 6. The BPNN was used as a benchmark to evaluate the RNN performance since the RNN was built using the Backpropagation algorithm.

For better comparison, the RNN used the same neural network parameters which were used by the BPNN classifier in Chapter 3.

The results obtained from training the RNN are shown in the table below.

**Table 15: Experimental results of RNN classifier trained using unpreprocessed dataset**

| None-RNN | | |
|---|---|---|
| k-Folds | Average Accuracy (%) | Average Training Time(s) |
| 15 | 85.7 | 27.85 |

The performance results of the BPNN and SVM can be seen in Table 5 and Table 10 respectively. The neural network parameters used in this experiment were not reconfigured but left the same throughout this work.

### 4.2.1.1 Results Discussion

The results of this experiment are compared to the experiments of the BPNN and SVM classifiers trained using the same dataset. The graph displayed in Figure 16 below shows the comparison of the results in a bar graph.

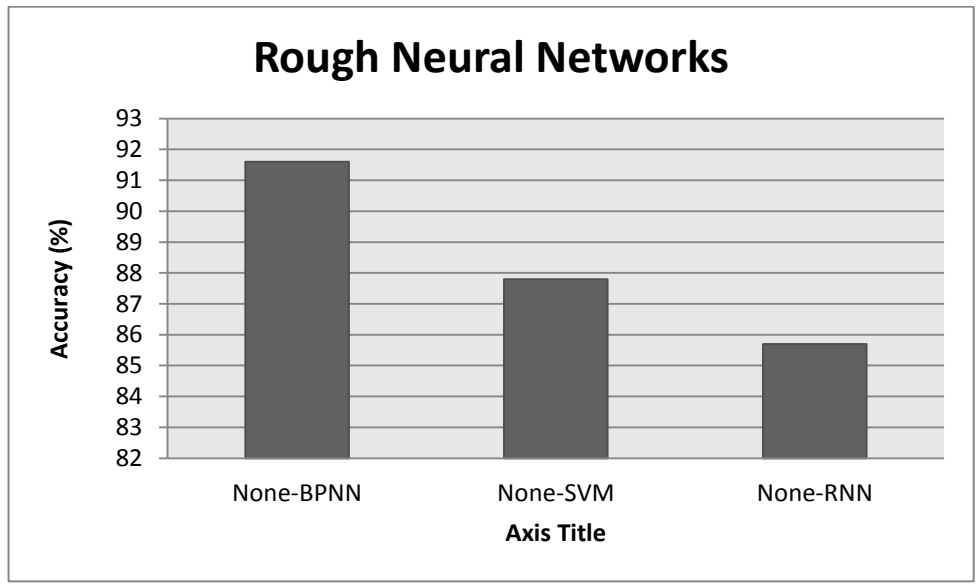

**Figure 16: Bar graph of classification accuracy of RNN, BPNN and SVM classifiers when trained with unpreprocessed dataset**

From the graph above it can be seen that the conventional BPNN and SVM classifiers triumph the newly introduced RNN classifier in this application. The evaluation of the classifier performance could also not be concluded solely on the classification accuracy. The time the classifiers took to learn was also used as an evaluation criterion, because if classifier was being adapted for online condition monitoring applications, the training time is an important aspect. The results are displayed in Figure 17 below.

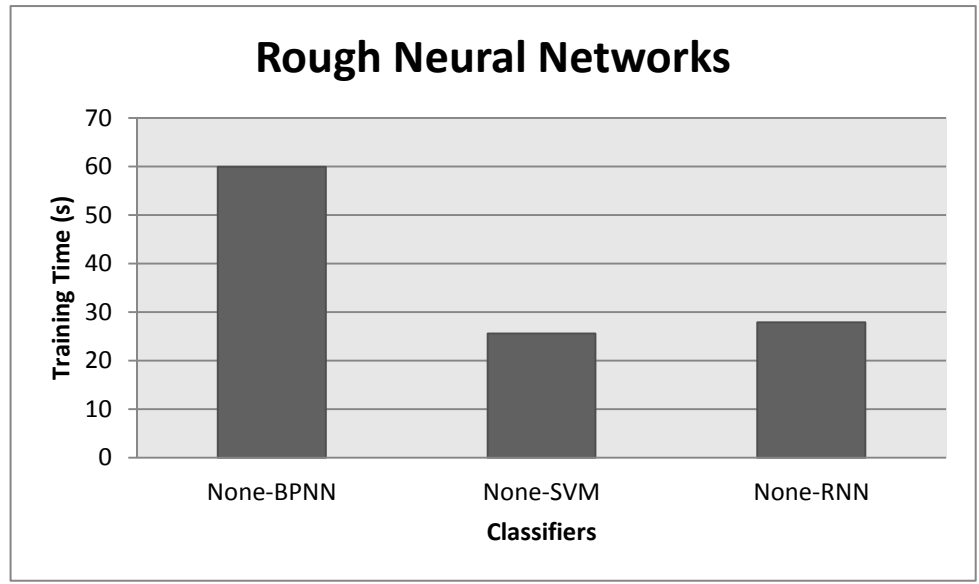

**Figure 17: Bar graph of average training time of RNN, BPNN and SVM classifiers when trained with unpreprocessed dataset**

In Figure 17 the RNN triumphs the BPNN learning time but could not surpass the SVM classifiers learning time. When the complete evaluation criteria are considered, it can be concluded that when the RNN trained using the raw unpreprocessed DGA data of the bushings faults it is outperformed by the BPNN and the SVM classifiers trained by the same dataset.

### *4.2.1.2 Conclusion*

This experiment investigated the performance of the RNN classifier in detecting faults in the bushings using DGA. The output results of this experiment were recorded on a table and then compared and evaluated against the performance of the BPNN and SVM classifiers.

Even though the RNN was built on the BPNN neural configurations, it didn't outperform it when raw unpreprocessed data was used for training. The next step in this chapter is the experiments that observe the effects that the preprocessors investigated on this work have on then RNN. This is discussed in the next section.

### 4.2.2 Training RNN Using Preprocessed Dataset

The last experiment in this work was to evaluate how data preprocessing affects the training of the bushings data when using RNN. The dataset obtained from the PCA, RS, GR++ and DT experiments is used to train the RNN.
The results obtained when RNN is trained using PCA are displayed in the table below.

Table 16: Experimental results of RNN classifier trained using principal component analysis preprocessed dataset

| PCA-RNN | | |
|---|---|---|
| k-Folds | Average Accuracy (%) | Average Training Time(s) |
| 15 | 99.7 | 15.91 |

The RNN was then trained using data obtained from the RS, where the following performance results were noted.

**Table 17: Experimental results of RNN classifier trained using rough sets preprocessed dataset**

| RS-RNN | | |
|---|---|---|
| k-Folds | Average Accuracy (%) | Average Training Time(s) |
| 15 | 90.3 | 10.23 |

The following results, Table 18 were obtained when RNN was trained using dataset obtained using the GR++ preprocessor.

**Table 18: Experimental results of RNN classifier trained using incremental granular ranking preprocessed dataset**

| GR++-RNN | | |
|---|---|---|
| k-Folds | Average Accuracy (%) | Average Training Time(s) |
| 15 | 61.1 | 5.63 |

The next part of the experiment was to train the RNN with the dataset obtained from the DT preprocessing experiments. The RNN trains one neural network with certain data and the other with uncertain data. During this experiment the neural network that is trained using uncertain data was not able to train, thus the only neural network that trained was similar to the one in the BPNN experiments.

Thus due to lack of uncertainty in the dataset obtained the classification performance of the RNN was not obtained when using DT dataset.

### *4.2.2.1 Results Discussion*

Experiment conducted in this section evaluates the effects of preprocessing DGA data used to train the RNN classifiers. The graph displayed in Figure 18 below shows the accuracy comparison of the RNN classifier when trained using dataset preprocessed by different methods.

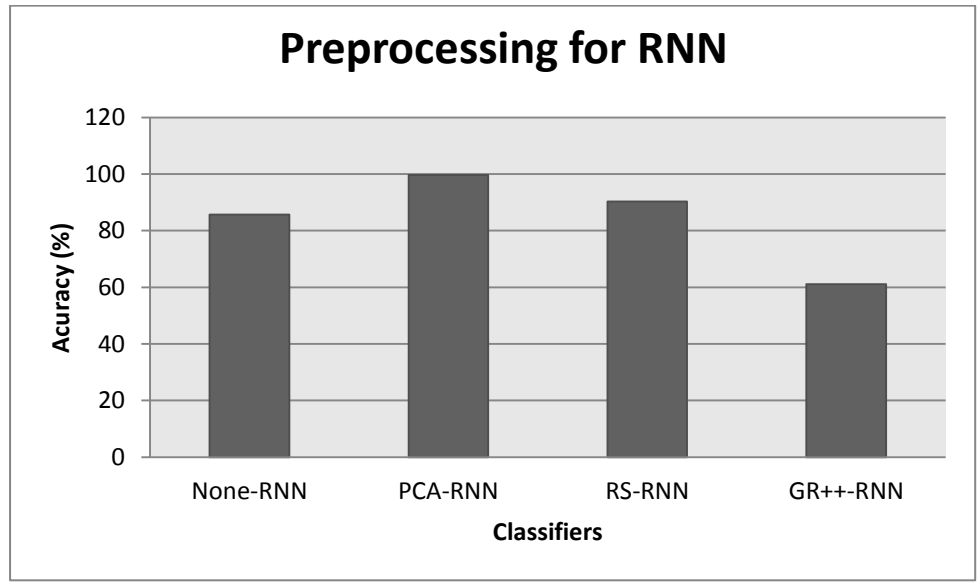

**Figure 18: Bar graph of classification accuracy of RNN classifiers when trained with different dataset**

The RNN's classification accuracy increased when trained with data with reduced attributes. The improvement can be seen when RNN is trained using dataset from the PCA preprocessing method and the ones from the RS method.

Classifier's accuracy when using PCA was 99.7% but when trained using GR++ preprocessed dataset, the classification accuracy decreased. The RNN performance was also evaluated using the training time criteria like the other classifiers investigated in Chapter 3. The results are shown in the graph in Figure 19.

All these preprocessors improve the training time of the RNN. This means that the PCA and RS preprocessors are good preprocessors for the RNN classifier because they improve its classification accuracy and training time. The GR++ is however a bad preprocessor because it reduced the accuracy of the classifier even though it managed to improve the training time.

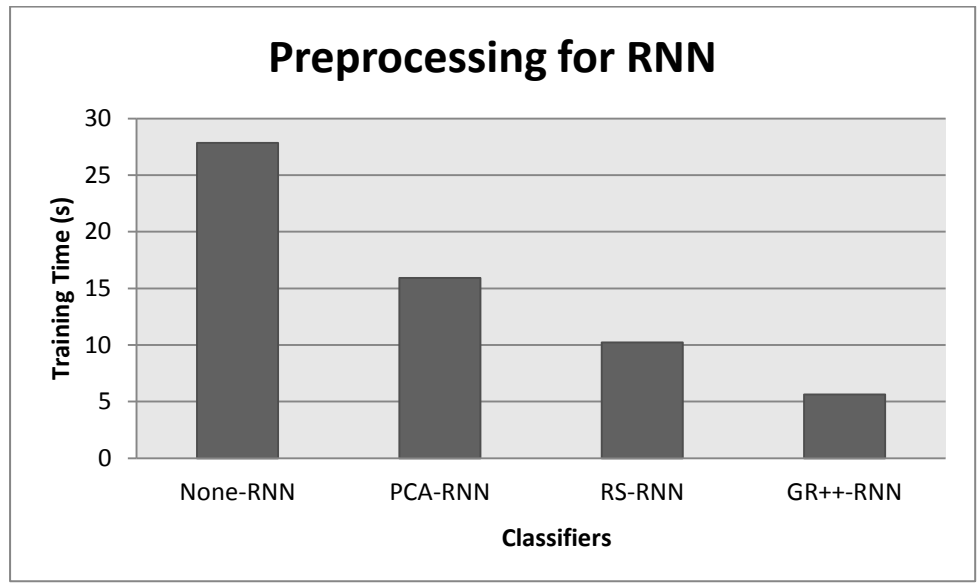

**Figure 19: Bar graph of average training time of RNN classifiers when trained with different dataset**

### *4.2.2.2 Conclusion*

From the experiments conducted in this section it is evident that the usages of PCA and RS preprocessors improve the performance of the RNN classifier. It means it is a good practice to reduce the attributes of the bushings DGA if the RNN is to be used for fault detection in bushings.

The dataset from the GR++ reduced the classification accuracy of the RNN classifier with a very big margin thus it is considered a bad preprocessor for RNN classifiers. The data from GR++ has the same effects as the effects it had on the previous experiments where it only increased the training time but compromised the classification accuracy.

### 4.3 Chapter Conclusion

This chapter investigated the RNN as a classifier of bushings faults data. The first experiments were to compare this new classifier to the classifiers that have been used in classification of bushings data which are BPNN and SVM. The results of the experiment showed that the BPNN and SVM classifiers outperformed the RNN classifier when raw unpreprocessed data was used.

The next part of the experiment was to evaluate the effects of the PCA, RS, GR++ and DT preprocessed data on the classification performance of the RNN. The experiment shows that data preprocessing has a positive effect training RNN classifiers since it improves the classifiers performance. The RNN classifier's training time improved when trained with the dataset from these classifiers, however the accuracy only improved when trained with the dataset from the RS and PCA.

When the RNN was used to classify the data from the DT experiments, the results showed that there were no uncertainties in the training dataset thus the experiment results were similar to the ones obtained when the BPNN was used.

The PCA had the highest effect compared to the effect that the RS had in improving the RNN classifier's performance. This chapter concludes that the PCA-RNN combination is better than the RS-RNN combination for this application and that one of these combinations must be considered if RNN classifier is considered as a classifier of choice in fault detection of DGA bushings. In the next chapter, the performance of the RNN with the preprocessors is compared to the performance of the combinations of BPNN and SVM classifiers with same preprocessors. This comparison is used to draw a conclusion on the best classifiers to use in this application.

# 5 OVERALL RESULT DISCUSSION AND CONCLUSION

## 5.1 Conclusion

This work searched for the most effective technique of DGA raw data dimension reduction (preprocessing) using the PCA, RS, GR++ and DT methods and how the preprocessing influenced the performance of BPNN and SVM classifiers. The work then further investigates the RNN classifier in this application. The performance of the RNN was evaluated against the performance of the BPNN and SVM when trained using unpreprocessed data. The effects of the preprocessors on the RNN classifiers were also investigated. In conclusion this chapter compares the performance of the RNN, BPNN and SVN classifiers' performance when both are trained using preprocessed data.

In this work the effects of reducing attributes in a multi-dimensional dataset used to train the classifiers for fault detection is used. To archive this objective, experiments were conducted using PCA, RS, GR++ and DT to reduce the attributes in the dataset. The resulting attributes showed that there exists a dependency and correlation relationship between ethane and total combustible gases in the decision making process for classifying transformer bushings. The presence of carbon dioxide also showed to have a strong influence in the decision making process since it was chosen by three out four preprocessors as final attributes. The presence of hydrogen gases was considered less influential despite the fact that they are present in most transformer fault types.

The SVM and BPNN classifiers were used to evaluate the effects that attribute reduction by each of these methods has in bushings fault detection applications. The obtained results showed that when the data is reduced using any of these preprocessors it can still be classified correctly.

RNN were also introduced for classification of uncertain dataset for bushings fault detection application using DGA dataset. This classifier proved to be worse than the BPNN and SVM classifiers, of which other researchers have also used for similar applications. Training the RNN with preprocessed dataset, however improved the performance. This performance results surpassed the performance of the BPNN and SVM classifiers when using the same dataset obtained using PCA and RS preprocessors.

The results obtained when classifiers were trained using PCA and RS preprocessed dataset indicate that attribute reduction has a positive effect in these applications. PCA improved the performance of both classifiers investigated in this work. GR++ constantly reduced the samples in the dataset which were used to train the classifiers. The reduction of samples is a property of the combination algorithm in the incremental ranking function. This property means that the final dataset has few samples, which are not enough to train the classifiers properly. Thus this led to poor classification in all classifiers. The reduction in the training time was also due to this property.

The resultant dataset from the Decision Tree pruning had no uncertain data, thus it can be concluded that the gases Carbon Dioxide, Ethane and the Total Combustible Gases are important gases in determining faulty transformers. The absence of uncertain data and high classification accuracy for the BPNN and SVM classifiers indicate that removal of the other gases did not compromise the quality of the dataset for decision making purposes. The absence of uncertainties was seen as influential when training the RNN classifier which requires the presence of uncertain data to learn.

GR++ dataset also caused RNN to have a lowest classification accuracy of 61.1% of which is about 20% less than the second lowest classification accuracy for the other preprocessor-classifier combinations. The poor performance of the RNN classifier when trained with GR++ dataset was because the RNN splits the data into upper and lower bounds, thus a small number of sample is passed to each lower and upper neuron of the RNN. This was however not the case when the BPNN was trained using the same GR++ dataset, where the performance was better than when RS dataset was used.

Preprocessing the dataset before training classifiers for bushings fault detection is a necessary step. If the RNN classifier is to be used in this application, the best suitable preprocessor should be determined to reduce the attributes of the dataset. From this work it can be concluded that the PCA-RNN combination produces the best classification performance results. This means that the "*curse of dimensionality*" problem was addressed and the uncertainties in the attribute reduced dataset was also addressed in one solution.

## 5.2 Suggestion of Future Work

- This work investigated the relations between the gases in DGA but did not investigate the kind of relationship that exists between each gas attribute. Determining the kind of relationship between the gases can pave a way for IEEE to address the statement they made in the IEEE C57-104 standard, which states that:
  > "*…it must be recognized that analysis of these gases and interpretation of their significance is at this time not a science, but an art subject to variability… The principal obstacle in the development of fault interpretation as an exact science **is the lack of positive correlation of the fault-identifying gases** with faults found in actual transformers.*"[58]

- The experiments in this work were focusing on fault detection; further work can be done to establish which gases needed for classifying faults to different fault types.
- The RNN classifier used in this work was based on the BPNN parameters for controlled test environment. Now that it has been established that it can make a good classifier, further investigations can be made where RNN will be modelled based on its own design parameters.